%% file: ms.tex
\pdfoutput=1 % Signal to arXiv to compile with PDFLaTex
\documentclass[runningheads]{llncs_eccv22}
\usepackage{graphicx}
\usepackage{tikz}
\usepackage{comment_eccv22}
\usepackage{amsmath,amssymb} % define this before the line numbering.
\usepackage{xcolor}
\usepackage{array,multirow,textcomp}
\usepackage{makecell}
\usepackage{csquotes}
\usepackage{epigraph}
\usepackage{adjustbox}

\RequirePackage[format=plain,labelformat=simple,labelsep=period,font=small,compatibility=false]{caption}
\RequirePackage[font=footnotesize,skip=3pt,subrefformat=parens]{subcaption}

\usepackage[accsupp]{axessibility}  % Improves PDF readability for those with disabilities.

\usepackage[width=122mm,left=12mm,paperwidth=146mm,height=193mm,top=12mm,paperheight=217mm]{geometry}

\usepackage{hhline}

\newcommand{\tableFont}{\fontsize{7pt}{8.4pt} \selectfont}

\definecolor{mygray}{gray}{.9}

\begin{document}
% \linenumbers
\pagestyle{headings}
\mainmatter
\def\ECCVSubNumber{2782}  % Insert your submission number here

\title{ Improving the Behaviour of Vision Transformers with Token-consistent Stochastic Layers}

%\title{Effective Stochasticity in Vision Transformers through Token-consistency} % Replace with your title

\begin{comment}
\titlerunning{ECCV-22 submission ID \ECCVSubNumber} 
\authorrunning{ECCV-22 submission ID \ECCVSubNumber} 
\author{Anonymous ECCV submission}
\institute{Paper ID \ECCVSubNumber}
\end{comment}
%******************

% For linking stuff from supplementary to the main paper
\def\FileIsMerged{}

%\begin{comment}
\titlerunning{ }
\author{
Nikola Popovic$^{1}$, Danda Pani Paudel$^{1}$, Thomas Probst$^{1}$, Luc Van Gool$^{1,2}$\\
}
\authorrunning{ }
\institute{Computer Vision Laboratory, ETH Zurich, Switzerland \and
VISICS, ESAT/PSI, KU Leuven, Belgium \\
\email{\{nipopovic, paudel, probstt, vangool\}@vision.ee.ethz.ch}\\ }
%\end{comment}
%******************
\maketitle

\input{latex/paper/00_abstract}
\input{latex/paper/01_introduction}

\input{latex/paper/02_related_work}

\input{latex/paper/03_background}
\input{latex/paper/04_method}
\input{latex/paper/05_experiments}
\input{latex/paper/06_conclusion}

%\clearpage

\appendix

\title{ \Large \bf
Improving the Behaviour of Vision Transformers with Token-consistent Stochastic Layers \\ Supplementary material
}
\titlerunning{ }
\author{
Nikola Popovic$^{1}$, Danda Pani Paudel$^{1}$, Thomas Probst$^{1}$, Luc Van Gool$^{1,2}$\\
%$^{1}$Computer Vision Laboratory, ETH Zurich, Switzerland\\
%$^{2}$VISICS, ESAT/PSI, KU Leuven, Belgium\\
% {\tt\small \{nipopovic, paudel, probstt, vangool\}@vision.ee.ethz.ch}
}
\institute{Computer Vision Laboratory, ETH Zurich, Switzerland \and
VISICS, ESAT/PSI, KU Leuven, Belgium \\
\email{\{nipopovic, paudel, probstt, vangool\}@vision.ee.ethz.ch}\\ }

\authorrunning{ }

\maketitle

\input{latex/supplementary/0_supplementary_structure}
\input{latex/supplementary/1_implementation_details}
\input{latex/supplementary/2_further_analysis}
\input{latex/supplementary/3_discussion}

\bibliographystyle{splncs04_eccv22}
\bibliography{references}
\end{document}

%% file: latex/paper/00_abstract.tex
\begin{abstract}
We introduce token-consistent stochastic layers in vision transformers, without causing any severe drop in performance. The added stochasticity improves network calibration, robustness and strengthens privacy. 
We use linear layers with token-consistent stochastic parameters inside the multilayer perceptron blocks, without altering the architecture of the transformer. 
The stochastic parameters are sampled from the uniform distribution, both during training and inference.
The applied linear operations preserve the topological structure, formed by the set of tokens passing through the shared multilayer perceptron. This operation encourages the learning of the recognition task to rely on the topological structures of the tokens, instead of their values, which in turn offers the desired robustness and privacy of the visual features. 
The effectiveness of the token-consistent stochasticity is demonstrated on three different applications, namely, network calibration, adversarial robustness, and feature privacy, by boosting the performance of the respective established baselines. 
%The source code will be made publicly available.   
%In this paper, we use our features for three different applications, namely, network calibration, adversarial robustness, and feature privacy.
% Our features offer exciting results on those tasks. 
%Furthermore, we showcase an experimental setup for federated and transfer learning, where the vision transformers with stochastic layers are again shown to be well behaved.
\end{abstract}

%% file: latex/paper/01_introduction.tex
\section{Introduction}
\label{sec:introduction}
Deep neural networks have shown to benefit from some level of stochasticity in various forms
%~\cite{srivastava2014dropout,shridhar2019comprehensive,spall2005introduction,panousis2021local,izmailov2018averaging,yu2019simple}.
%the form of dropout~\cite{srivastava2014dropout}, variational bayesian inference~\cite{shridhar2019comprehensive}, and other techiques~\cite{spall2005introduction,panousis2021local,izmailov2018averaging,yu2019simple}. 
%Stochasticity can in fact serve a variety of purposes, 
including regularization~\cite{srivastava2014dropout}, optimization ~\cite{spall2005introduction,izmailov2018averaging,Smith2021OnTO}, data augmentation~\cite{Cubuk2020RandaugmentPA}, as well as monte-carlo sampling for variational inference \cite{Kingma2014AutoEncodingVB,Gal2015BayesianCN,shridhar2019comprehensive,yu2019simple} or uncertainty estimation~\cite{Gal2015BayesianCN,Gal2016DropoutAA,Kendall2017WhatUD}.
%, as well as gradient obfuscation for adversarial robustness~\cite{Rakin2019ParametricNI,panousis2021local}.
In general, it is however not clear how much and what form of stochasticity is desired for given neural architectures and for the tasks to be performed. 

In this work, we introduce and systematically analyze the effect of token-consistent stochastic layers in vision transformers (ViTs)~\cite{dosovitskiy2021image}.
ViTs are networks of gripping interest, not only due to their widespread use but also due to their versatility. 
In turn, transformer architectures allow us to introduce stochastic layers in a controlled and tractable manner. 
This is achieved by preforming token-consistent stochastic operations inside each block of the transformer, without altering the architecture or introducing additional learnable parameters. 
It is only necessary to briefly fine-tune the ViT, in order to adapt it to the stochasticity.
These stochastic layers are introduced in an attempt to render the visual features more robust, private and suitable for good calibration, while maintaining the recognition performance of the initial visual transformers.

%These stochastic operations are performed on each channel dimension of every feature, thus making the layers fully stochastic in nature.    

%\input{figures/paper/main_teaser}

One important aspect of our way of injecting stochasticity is with regard to its consistency across the tokens. 
More precisely, we perform the stochastic linear operation with the same sample on all token features passing through a given multilayer perceptron (MLP), with a completely non-informative uniform distribution as the stochasticity source.
Such token-consistent linear operations preserve the topological structures, invariant under affine transformations. Therefore, our way of injecting stochasticity encourages  recognition to rely on the topological structures formed by the tokens, instead of their values.
We demonstrate the effectiveness of token-consistent stochastic layers through improvements on the calibration and robustness of the vision transformers.
The improved behaviours are demonstrated by comparing against the existing methods and the established baselines.
Furthermore, the visual features are more private, when shared, because it is more difficult to reconstruct the input image due to the preformed stochastic operations.
%At the same time, the stochastic operations on the visual features makes them more private, when shared, as the recovery of the previous stage is particularly difficult due to the unknown stochastic operations.
In this process, the privacy is achieved without severe impacts on image recognition.
%Experiments show that the private feature extractor still preserves the information necessary to perform a variety of vision tasks. 
Our process of stochasticity injection tailored to vision transformers differs from existing ones ~\cite{srivastava2014dropout,shridhar2019comprehensive,panousis2021local,yu2019simple} in one or more of these aspects: computational, tractability, or performance. The details are discussed in Section~\ref{sec:background},~\ref{sec:stochastic_layers_VT}~and~\ref{sec:experiments}. 

\renewcommand{\labelitemi}{$\bullet$}
The key contributions of this work are:
\vspace{-5pt}
\begin{itemize}
    \item We propose a token-consistent stochastic layer, which can be inserted in standard vision transformers without modifying the original architecture.
    \item We study the behaviour of the proposed stochasticity through exhaustive experiments on the ImageNet~\cite{Russakovsky2015ImageNet} benchmark dataset.
    \item We demonstrate its effectiveness on network calibration, adversarial robustness, and feature privacy, without severe impacts on image recognition.
    %We demonstrate no severe performance degradation due to the proposed stochasticity, while improving the network calibration, adversarial robustness, and feature privacy.
    %that completely non-informative stochasticity source
    %\item Observing no damage to the performance, while using a completely non-informative stochasticity source. Furthermore, observing that token-consistent noise is stronger than sampling different noise for different tokens.
  %  \item Interpretation of keeping the topological structure of features during prediction.
\end{itemize}

% In this paper, we examine our stochastic layers with regards to adversarial robustness, network calibration, and feature privacy. Our features offer exciting results for those. Furthermore, we also showcase experimental setups for federated and transfer learning, where the vision transformers with stochastic layers are again shown to be well behaved. 

%% file: latex/paper/02_related_work.tex
\section{Related Work}
\label{sec:related_work}

% In what follows, we present related work with regards to the transformer architecture and stochasticity on a high level. For a detailed discussion of the literature on the issues of adversarial robustness, private federated learning, and network calibration, please refer to Section~\ref{sec:experiments}.

\textbf{Vision Transformer.}
The transformer was first introduced in natural language processing~\cite{vaswani2017AllYouNeed,dai2019transformer,devlin2019bert,yang2019xlnet}, based on the self-attention mechanism~\cite{vaswani2017AllYouNeed}. 
In order to capture context across input sequences, the transformer employs multi-head self-attention and multi-layer perceptron modules, allowing for global interaction. 
Through the seminal work of ViT~\cite{dosovitskiy2021image}, transformers have successfully entered the vision domain.
%by dividing an image into patches that are treated as a sequence. 
Vision transformers achieve state-of-the-art performance on problems including image classification~\cite{liu2021swin,yuan2021tokens,wang2021pyramid}, object detection~\cite{carion2020end}, semantic/instance segmentation~\cite{SETR}, and video segmentation~\cite{wang2020end}. For a more complete survey of vision transformers, we refer to \cite{khan2021transformers}.

\noindent\textbf{Stochasticity in Neural Networks.}
Stochasticity has been a subject of study with regards to its effects on regularization and optimization~\cite{srivastava2014dropout,Neelakantan2015AddingGN,Smith2021OnTO}, data augmentation~\cite{Cubuk2020RandaugmentPA,LingChen2020UniformAugmentAS}, monte-carlo sampling for variational inference of latent variables~\cite{Kingma2014AutoEncodingVB} and model parameters~\cite{Gal2015BayesianCN,Kingma2015VariationalDA}, as well as generative modelling~\cite{Kingma2014AutoEncodingVB,Mordido2018DropoutGANLF,Karras2019ASG}. 
Furthermore, noise injection has been successfully applied to uncertainty estimation and network calibration~\cite{Gal2015BayesianCN,Gal2016DropoutAA,Kendall2017WhatUD,Laves2020CalibrationOM} and adversarial robustness~\cite{Alemi2017DeepVI,Rakin2019ParametricNI,Cohen2019CertifiedAR}, and network compression~\cite{Dai2018CompressingNN}.

\noindent\textbf{Confidence calibration.} 
In real-world decision making systems, classification networks should also indicate when they are likely to be incorrect~\cite{Naeini2015ObtainingWC,guo2017calibration}. While it was shown early on that shallow neural networks typically produce well-calibrated probabilities~\cite{Niculescu2005goodProb}, it was discovered later that modern networks are not well-calibrated~\cite{guo2017calibration}. 
MC-Dropout~\cite{Gal2016DropoutAA,srivastava2014dropout} and Deep Ensambles~\cite{lakshminarayanan2017ensambleUncertanity} are the most popular techniques for combating this issue. A more structured way to drop model parameters is also presented as a good solution~\cite{Durasov21Masksembles}. 
A thorough analysis of the calibration level of vision transformer architectures can be found at~\cite{Minderer2021RevisitingTC}.
%When it comes to transformers, it has been shown that temperature scaling is effective at further reducing calibration errors in NLP~\cite{Desai2020TransfCalib}. For a thorough analysis the calibration level of various vision transformer architectures, take a look at~\cite{Minderer2021RevisitingTC}.

\noindent\textbf{Adversarial robustness.}
A small adversarial perturbation in pixel intensities can lead to a severe drop in performance of accurate deep networks~\cite{Szegedy2014IntriguingPO,Goodfellow2015ExplainingAH}. Most of the popular adversary attacks exploit the networks differentiability \cite{Szegedy2014IntriguingPO,Goodfellow2015ExplainingAH,Madry2018TowardsDL,Athalye2018SynthesizingRA,Kurakin2016AdvPhysical,mihajlovic2018CommonAdv}.
Numerous approaches were proposed to defend against these attacks, which we roughly categorize into adversarial training~\cite{Kurakin2017AdversarialML,Shafahi2019AdversarialTF,Wong2020FastIB,Madry2018TowardsDL,Zhang2019TheoreticallyPT}, certified defenses~\cite{Cohen2019CertifiedAR,Croce2019ProvableRO,Zhai2020MACERAA,Salman2019ProvablyRD}, gradient regularization~\cite{Ciss2017ParsevalNI,Ross2018ImprovingTA,Finlay2019ScaleableIG,Ororbia2016UnifyingAT,Jakubovitz2018ImprovingDR,Gu2015TowardsDN}, and stochasticity injection~\cite{Rakin2019ParametricNI,46641,Pinot2019TheoreticalEF,Lee2021GradDivAR}. Furthermore, it is interesting that adversarially robust networks have other beneficial properties~\cite{salman2020adveTransfer,Ilyas2019AdversarialEA,kaur2019PerceptuallyAligned}.
%
% In particular,~\cite{Pinot2019TheoreticalEF,Rakin2019ParametricNI,Lee2021GradDivAR} show that injection of stochasticity into the network improve adversarial robustness. In that line, our goal is to investigate the impact of our stochastic layers with regards to the adversarial robustness of the network.

\noindent\textbf{Feature privacy.}
Sharing image features is an important part of collaborative distributed systems~\cite{Singh2021DISCODA,Roy2019MitigatingIL,Li2021DepObfuscation}.
Applications include camera localization~\cite{Speciale2019PrivacyPI}, structure-from-motion~\cite{Geppert2020PrivacyPS}, federated learning~\cite{bonawitz2019towards,kairouz2019advances,Konecn2016FederatedLS}, or split learning/inference \cite{gupta2018distributed,vepakomma2018split}.
To address privacy issues,
%, problem-specific solutions have been proposed.
\cite{Pittaluga2019RevealingSB} obfuscates the geometry,
%to maintain privacy of sparse features~\cite{Pittaluga2019RevealingSB}. 
homomorphic encryption guarantees privacy of feature extraction~\cite{Qin2014TowardsEP,Hsu2012ImageFE}, retrieval~\cite{Engelsma2020HERSHE}, and federated learning~\cite{Fang2021PrivacyPM}. 
However, these methods do not readily extend to many practical cases of distributed learning, like split learning~\cite{gupta2018distributed,vepakomma2018split}.
%where features of the data are shared, after being processed by a part of the complete network.
Even though the feature extraction removes information from the input image, it is still possible reconstruct the input~\cite{Ulyanov2018DeepIP,Zhang2020TheSR,Yin2020DreamingTD,Wang2021IMAGINEIS} or private attributes~\cite{Basu2019MembershipMI,Singh2021DISCODA}.
%The local network could be a generic visual feature extractor that may also be trained. The top part of the network on the other hand is then processed on a centralized server, without sharing the original images. In similar fashion, one could split the workload for inference across different devices, e.g. across a mobile device and a server.

%% file: latex/paper/03_background.tex
% \epigraph{If people do not believe that mathematics is simple, it is only because they do not realize how complicated life is.}{\emph{John von Neumann (1903-1957)}}{\vspace{-3mm}}

\section{Background}
\label{sec:background}

The goal of introducing stochasticity by noise injection is to teach a network to (not) focus on specific aspects of the data. 
%For instance, Dropout~\cite{srivastava2014dropout} aims to mitigate co-adaption of neurons, to prevent overfitting to only a few neural pathways. 
We now present our hypothesis that the injection of noise into a transformer architecture, under certain conditions, guides the network to rely on topological features.

\noindent\textbf{General Topology:} Given a set of $n$ points, $\mathcal{X} = \{\mathbf{x}_i\}_{i=1}^n \in \mathbb{R}^d$ assumed to be sampled on or near some underlying topological space $X \! \subseteq \!\mathbb{R}^d$, we are interested in the information about the topology of $X$. To be precise, we are interested to perform a stochastic modification of the point set $\mathcal{X}$ such that the underlying topology of $X$ remains unchanged in some sense. 
Informally two topological spaces $X$ and $Y$ are equivalent ($X \cong Y$) if there is a continuous function $f\!: X \! \rightarrow \! Y$ that has an inverse $f^{-1}$ which is also continuous. 
%\PR{we have to be careful with bernoulli dropout here.}
Whenever the mapping $f$ is feasible,  we call these two topological spaces $X$ and $Y$ homeomorphic and $f$ is their homeomorphism. For a more detailed and formal account on topology please refer to~\cite{guss2018characterizing,chazal2021introduction}.

\begin{figure}[t]
\centering
\includegraphics[width=0.5\columnwidth]{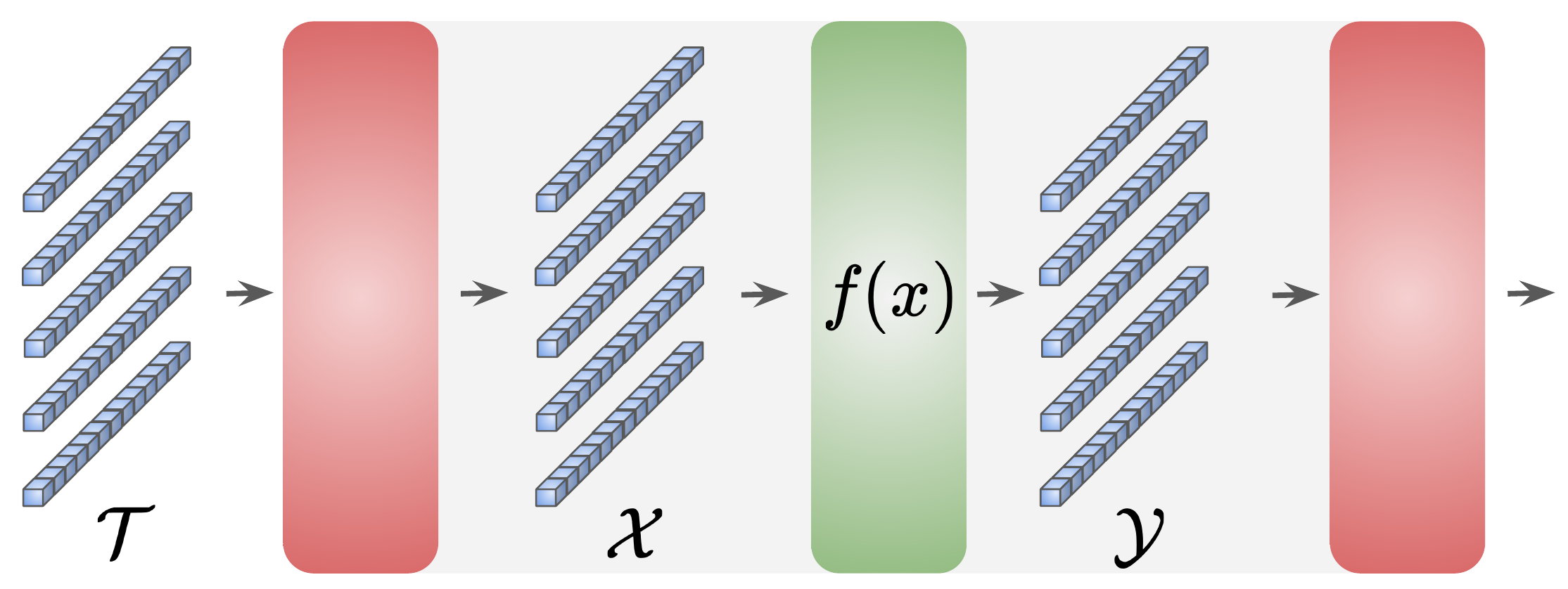}
\vspace{-4pt}
\caption{A single stochastic multilayer perceptron with multiple input tokens $\mathcal{T}$. At the intermediate stage, the stochastic layer $f(\mathbf{x})$  homeomorphically maps features $\mathcal{X}$ to $\mathcal{Y}$.}
\label{fig:slayBlock}
\end{figure}

In this work, the set of points $\mathcal{X}$ are the features generated by $n$ tokens of the transformers. We wish to encourage the recognition task to rely upon the topology of $X$, instead of the entries of the feature vectors. \emph{Therefore, we would like to preserve the topology of $X$ during the stochastic operations}. To do so, we perform a stocastically homeomorphic mapping of the point set $\mathcal{X}$. In our experiments, we use an invertible linear transformation  --around the identity map for numerical stability-- to represent $f$, which is by design a homeomorphic mapping. A graphical illustration of the performed operations is shown in Figure~\ref{fig:slayBlock}.

% \def\svgwidth{0.5\textwidth}
%  %\input{images/paper/slay.pdf_tex}
% ------------------- CVPR 2 column settings --------------------
% \centering
% \adjustbox{trim=0.25cm 1.4cm 0.5cm 0.75cm, clip}{%
% \includegraphics[width=\columnwidth]{images/paper/mlp.pdf}
% }
% ---------------------------------------------------------------

%\vspace{-3pt}
%\vspace{-3pt}

\noindent\textbf{Inside the multilayer perceptron in ViTs:}
ViTs divide images into multiple tokens. These tokens are processed by attention modules followed by multilayer perceptrons (MLPs). We are interested in introducing stochasticity within the processing pipeline of the MLPs. Let us consider, without lack of generality, that the intermediate features inside an MLP for tokens $\mathcal{T} = \{t_i\}_{i=1}^n$  are given by
$\mathcal{X} = \{\mathbf{x}_i\}_{i=1}^n \in \mathbb{R}^d$.  A homeomorphic mapping $f:X \rightarrow Y$, in a fully stochastic fashion, is applied to the set of features $\mathcal{X} $ resulting into the mapped set $\mathcal{Y}$. We continue processing the mapped features $\mathcal{Y}$ similar to $\mathcal{X}$, as if the stochastic layer represented by $f$ was absent. We represent $f$ as a square diagonal matrix $\mathbf{A}\in\mathbb{R}^{d\times d}$ with non-zero diagonal entries.  Then the operation by the introduced stochastic layer in the form, $\mathbf{y}_i = f(\mathbf{A},\mathbf{x}_i)= \mathbf{Ax}_i, \forall \mathbf{x}_i\in\mathcal{X}$, ensures $f:X\rightarrow Y$ to be homeomorphic. Please note that the mappings by square diagonal matrices with non-zero entries are 
continuous with their inverse being continuous as well. Other choices of $f$ could be made as well, but we study the behaviour of ViTs with a relatively simple function $f$. 

%% file: latex/paper/04_method.tex
\section{Token-consistent Stochastic Layers in Vision Transformers}
\label{sec:stochastic_layers_VT}
%We now describe in detail how our stochastic layer is employed inside a vision transformer.

Let the input image be denoted as $\mathcal{I} \in \mathbb{R}^{H \times W \times 3}$, where $H$ is the height and $W$ is the width of the image. 
The image is represented using $3$ color channels. 
A feature extractor $\mathsf{F}$ takes the image $\mathcal{I}$ as the input and produces a feature map $\mathcal{F} = \mathsf{F} (\mathcal{I})$.
Usually $\mathcal{F} \in \mathbb{R}^{\frac{H}{s} \times \frac{W}{s} \times C}$, meaning that its spatial dimensions are downscaled by a factor of $s$ compared to the image $\mathcal{I}$ and that it has $C$ channels. 
The feature map $\mathcal{F}$ can be used for various tasks.
As one option, the feature map can be an input to a classification model $\mathsf{C}$ to produce $\mathcal{O} = \mathsf{C} (\mathcal{F}) = \mathsf{C} (\mathsf{F}(\mathcal{I}))$, where $\mathcal{O} \in \mathbb{R}^{m}$ and $m$ is the number of classes. 
It can also be given to a regression model $\mathsf{R}$ to produce $\mathcal{O} = \mathsf{R} (\mathcal{F})$, where $\mathcal{O} \in \mathbb{R}^{m}$ and $m$ is the number of regressed values.
Another option is to give it to a model $\mathsf{D}$ to produce  an image-to-image translation output $\mathcal{O} = \mathsf{D} (\mathcal{F})$, where $\mathcal{O} \in \mathbb{R}^{H \times W \times m}$. 
This can be used for various tasks like semantic segmentation, depth estimation, flow estimation, etc.

Generally speaking, a Vision Transformer~(ViT)~\cite{dosovitskiy2021image} is one type of feature extractor $\mathsf{F}$. 
First, it takes the input image $\mathcal{I}$ and divides it spatially into $k \times k$ patches to obtain $\mathcal{\Tilde{I}} \in \mathbb{R}^{\frac{H}{k}\frac{W}{k} \times 3k^2}$, which represents $n_{T}=\frac{H}{k} \frac{W}{k}$ token vectors of dimension $3k^2$. 
A popular choice is $k=16$~\cite{dosovitskiy2021image,Touvron21aDeIT}. 
Then, each token vector is embedded into $d_{T}$ dimensions, using the same affine transformation, to obtain $\mathcal{F}_0 \in \mathbb{R}^{n_{T} \times d_{T}}$. 
Optionally, a classification or regression token can be appended to $\mathcal{F}_0$, which would result in $\hat{n_{T}}=\frac{H}{k} \frac{W}{k} + 1$. 
This token is used to make a prediction for the final task.
Next, a positional encoding $\mathcal{P} \in \mathbb{R}^{n_{T} \times d_{T}}$ is added to $\mathcal{F}_0$ which gives $\mathcal{\Tilde{F}}_0 = \mathcal{F}_0 + \mathcal{P}$. 
The purpose of the positional encoding $\mathcal{P}$ is to add information about which token in $\mathcal{F}_0$ corresponds to which spatial location of the input image $\mathcal{I}$.
The transformer then uses $l$ consecutive transformer blocks $\mathsf{B}_i$ to extract features. 
Each block $\mathsf{B}_i$ takes the output of the previous block $\mathcal{F}_{i-1}$ as its input and produces $\mathcal{F}_i = \mathsf{B}_i(\mathcal{F}_{i-1})$, where  $\mathcal{F}_{i} \in \mathbb{R}^{n_{T} \times d_{T}}$. 
The first block takes $\mathcal{\Tilde{F}}_{0}$ as input. 
To obtain the final feature map $\mathcal{F}$, we usually take the output of the last block $\mathcal{F}_{l}$, discard the optional classification or regression token, and spatially rearrange it back as $\mathcal{F} \in \mathbb{R}^{\frac{H}{s} \times \frac{W}{s} \times C}$, where $C=d$ and $s=k$.

Each transformer block $\mathsf{B}_i$ is composed of a Multi-head Self Attention block (MSA), followed by a Multilayer Perceptron (MLP)~\cite{vaswani2017AllYouNeed,dosovitskiy2021image}.
Self-attention is a spatially global operation, where every token interacts with every other token and thus information is shared across spatial dimensions.
Multiple heads in the MSA block are used for more efficient computation and for extracting more diverse features.
The MSA block executes the following operations:
\begin{equation}
    \label{eq:MSA}
    \mathcal{Z}_{i} = 
    \mathsf{MSA}(\mathsf{LN}(\mathcal{F}_{i-1})) + 
    \mathcal{F}_{i-1},
\end{equation}
where $\mathsf{LN}$ represents layer normalization~\cite{ba2016LayerNorm}.
The MSA block is followed by the MLP block, which processes each token separately using the same multilayer perceptron. 
This block processes the tokens, after their spatially global interactions in the MSA block, by sharing and refining the representations of each token across all of its channels. 
The MLP block executes the following operations:
\begin{equation}
    \label{eq:MLP}
    \mathcal{F}_{i} = 
    \mathsf{MLP}(\mathsf{LN}(\mathcal{Z}_{i})) + 
    \mathcal{Z}_{i},  \quad \mathcal{Z}_{i} \in \mathbb{R}^{n_{T} \times d_{T}}.
\end{equation}
In the following, we omit the block index $i$ for simplicity,
and unpack~\eqref{eq:MLP}:
%Unpacking~\eqref{eq:MLP} corresponds to the operations:
\begin{alignat}{2}
    \label{eq:MLP_details}
    %& &\mathcal{Z}_{i} \in \mathbb{R}^{n_{T} \times d}, \\
    &\mathcal{Z}^{LN} = \mathsf{LN}(\mathcal{Z}),&& 
    \mathcal{Z}^{LN} \in \mathbb{R}^{n_{T} \times d_{T}}, \\
    &\mathcal{Z}^{FC_1} = \mathsf{FC}(\mathcal{Z}^{LN}),&& 
    \mathcal{Z}^{FC_1} \in \mathbb{R}^{n_{T} \times d},\\
    \label{eq:MLP_details_act}
    &\mathcal{Z}^{act} = \sigma(\mathcal{Z}^{FC_1}),\qquad && 
    \mathcal{Z}^{act} \in \mathbb{R}^{n_{T} \times d},\\
    &\mathcal{Z}^{FC_2} = \mathsf{FC}(\mathcal{Z}^{act}),&& 
    \mathcal{Z}^{FC_2} \in \mathbb{R}^{n_{T} \times d_{T}},\\
    &\mathcal{F} = \mathcal{Z}^{FC_2} + \mathcal{Z},&&
    \mathcal{F} \in \mathbb{R}^{n_{T} \times d_{T}},
\end{alignat}
%with fully connected layers $\mathsf{FC}$ and activation function $\sigma$.
where $\mathsf{FC}$ is the fully connected layer and $\sigma$ the activation function.

\begin{figure}[t!]
\centering
\includegraphics[width=0.6\columnwidth]{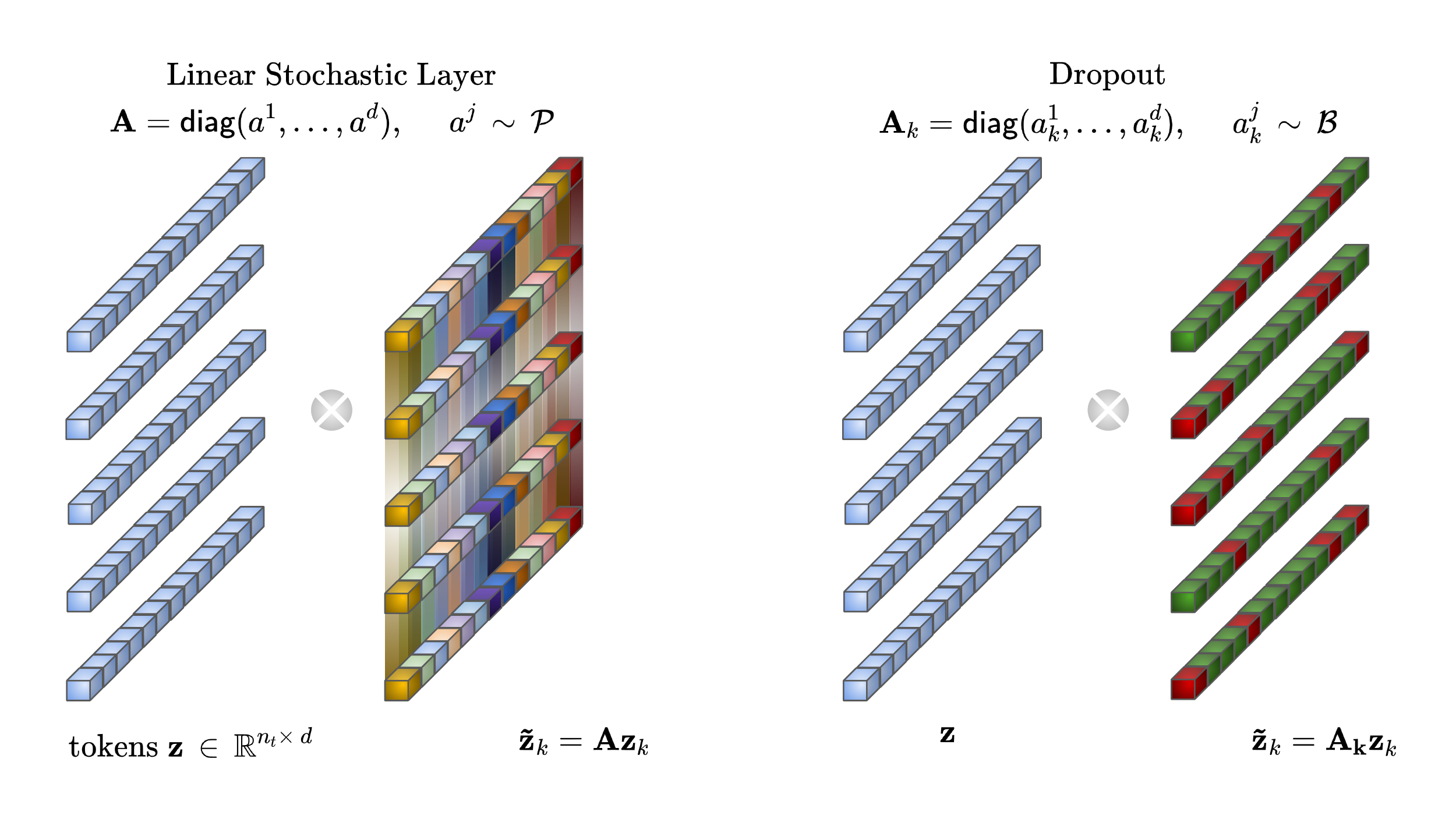}
\vspace{-9pt}
\caption{Token-consistent Stochastic Layer compared to dropout~\cite{srivastava2014dropout} regularization, illustrated with the help of vision transformer's tokens.}
\label{fig:SLAvsDO}
\end{figure}

We inject noise into the transformer, by injecting noise into the intermediate feature map $\mathcal{Z}^{FC_1} \in \mathbb{R}^{n_{T} \times d}$ of every MLP.
%We modify every transformer block by injecting noise into the intermediate feature map $\mathcal{Z}^{FC_1} \in \mathbb{R}^{n_{T} \times d}$ of the MLP~\eqref{eq:MLP_details}. 
More specifically, this is achieved by multiplying all of its tokens $\mathbf{z}^{FC_1}$ with a stochastic square diagonal matrix $\mathbf{A}\in \mathbb{R}^{d \times d}$ to obtain $\mathbf{\Tilde{z}}^{FC_1} \in \mathbb{R}^{d}$, and thus $\mathcal{\Tilde{Z}}^{FC_1} \in \mathbb{R}^{n_{T} \times d}$ as,
\begin{equation}
    \label{eq:noise_injection}
    \begin{split}
        &\mathbf{\Tilde{z}}^{FC_1}_k = 
        \mathbf{A}\mathbf{z}^{FC_1}_k, \quad k = 1\dots n_{T}\\
        &\mathbf{A} = \mathsf{diag}(a^{1}, a^{2}, ...., a^{d}), \quad a^{j} \sim \mathcal{P}.%\\
      %  &a^{j} \sim \mathcal{P}.
    \end{split}
\end{equation}
This corresponds to multiplying the channel dimension $j$ of all tokens with the same $a^{j}$ (token-consistent). 
The elements of $\mathbf{A}$ are sampled from a continuous random distribution $\mathcal{P}$, independently for each block $\mathsf{B}_i$.
The matrix $\mathbf{A}$ is sampled only once per input in $\mathsf{B}_i$, meaning that each token's feature vector $\mathbf{\Tilde{z}}^{FC_1}$ gets multiplied element-wise with the same noise vector, akin to a stochastic linear layer. The feature map $\mathcal{\Tilde{Z}}^{FC_1}$ is used instead of $\mathcal{Z}^{FC_1}$ in~\eqref{eq:MLP_details_act}.
We choose to inject the noise to the feature map $\mathcal{Z}^{FC_1}$, because it has the highest dimensionallity in the block ($d > d_T$), thus the impact of the noise will be the highest.
If $\mathbf{A}_k$ was re-sampled for different tokens $k$, and $\mathcal{P}$ was the Bernoulli distribution, this would correspond to the commonly known dropout regularization~\cite{srivastava2014dropout}. Figure~\ref{fig:SLAvsDO} highlights the difference between our method and dropout visually.

% ------------------- CVPR 2 column settings --------------------
% \includegraphics[width=\columnwidth]{images/paper/sla_vs_dropout.pdf}
% ---------------------------------------------------------------

%% file: latex/paper/05_experiments.tex
\section{Experiments}
\label{sec:experiments}
We now empirically study the the effectiveness of token-consistent stochastic layers, by gauging their impact on confidence calibration and robustness of the vision transformers, under different strength of stochasticity. At the same time, we analyse the trade-offs with respect to classification accuracy, privacy and transferability of the visual features. 
For each of these aspects, we compare our method to established baselines and applicable state-of-the-art alternatives, in order to demonstrate its impact in a holistic manner. For additional results, more detailed discussions and a further analysis with regards to topological properties, we would like to refer to our supplementary materials.

We choose the DeiT-S architecture~\cite{Touvron21aDeIT} as our main representative of vision transformers in our experiments. DeiT-S is one specific instance of the ViT ~\cite{dosovitskiy2021image} and has a parameter count and computational complexity similar to  ResNet-50~\cite{He2015ResNet}. When analysing the effect of our stochastic layers on the ImageNet accuracy in Section~\ref{sec:experiments_stochastic_layers}, we also verify that the behaviour is consistent across vision transformer architectures.
After asserting a tenable decrease in accuracy, we continue to evaluate the suitability of the proposed stochastic layers with regards to confidence calibration in Section~\ref{ssec:confidence_calibration}, and defending adversarial attacks Section~\ref{sec:adversarial_robustness}.
To round up the empirical analysis, we further investigate the impact visual features with regards to privacy in collaborative settings, and their ability to generalize to new tasks in Sections~\ref{ssec:private_features}
and~\ref{sec:transfer_learning}, respectively.

% In this section, we analyze the effect of our token-consistent stochastic layers for vision transformers on several aspects, including classification \textbf{accuracy}, confidence \textbf{calibration}, adversarial \textbf{robustness}, network \textbf{privacy}, and \textbf{transfer} learning, considering several image understanding tasks.
% For this purpose, we will use the DeiT-S architecture~\cite{Touvron21aDeIT}. DeiT-S is one specific instance of the ViT transformer~\cite{dosovitskiy2021image} and has a parameter count and computational complexity similar to  ResNet-50~\cite{He2015ResNet}.
%\textcolor{red}{DANDA: I am back here. Maybe we need make everything  clear here itself with proper highlights regarding the experiments: why are we doing what we are doing.. }

%%%%%%%%%%%%%%%%%%%%%%%%%%%%%%%%%%%%%%%%%%%%%%%%%%%%%%%%%%%%%%%%%%%%%%%%%%%%%%%%%%%%%%%%%%
\subsection{Effects on ImageNet Accuracy}
\label{sec:experiments_stochastic_layers}

\subsubsection{Training procedure.} 
To explore the general behaviour of our method we use the ILSVRC-2012 ImageNet-1k dataset, which contains $1.2$M training and $50000$ validation images grouped into $1000$ classes~\cite{Russakovsky2015ImageNet}. The DeiT-S model pre-trained on ImageNet-1k is used as a starting point for all experiments in Section~\ref{sec:experiments}, unless otherwise mentioned. All the details regarding the pre-training procedure are contained in~\cite{Touvron21aDeIT}. We insert the token-consistent stochastic layer from \eqref{eq:noise_injection} into every block of the DeiT-S and fine-tune it for $30$ epochs. We choose $\mathcal{P}$ to be uniform, $\mathcal{P} \sim \mathcal{U}(1-\Delta,1+\Delta)$, since it is a completely non-informative distribution and thus the strongest source of noise. During this fine-tuning process, we use the AdamW optimizer~\cite{Loshchilov2017AdamW} and a cosine scheduler with a learning rate of $10^{-5}$, which gradually decays to $10^{-6}$. We also turn off DropPath~\cite{huang2016stochasticDepth} and use the ReLU activation in order to introduce the noise to a deterministic baseline with a (piecewise) linear impact.
%We fine-tune a completely deterministic baseline as well. 
All other settings remain as in pre-training. 

\input{figures/paper/classification}

\subsubsection{Experiments.} 
We analyzing the predictive performance of stochastic layers on the ImageNet validation set, measured through classification accuracy. As presented in Figure~\ref{fig:classification_accuracy}, there is no significant drop in accuracy, even when using high noise levels during inference. 
The network performs better when $a^{j}$ from ~\eqref{eq:noise_injection} is set as the mean of the training noise distribution $\mathbb{E}\left[\mathcal{P}\right]$ during inference (noise turned off). Also, the network preforms better when multiple predictions are sampled ($a^{j}\sim\mathcal{P}$ independently sampled for each block and prediction) and averaged in a Monte Carlo fashion. However, the inference also works reasonably well when we just sample once. This is surprising given the fact that $a^{j}$, which multiplies token channels, is sampled from a completely non-informative uniform distribution.
In Table~\ref{table:imnet_1k_accuracy_thorough}, we see that our stochastic layer preforms slightly worse than regular dropout, indicating that it is a stronger source of stochasticity. When analyzing a certain $\Delta$, we chose the dropout probability such that it has the same mean and variance.  Also, when the noise matrix $\mathbf{A}$ is sampled independently for each token (non-token-consistent), the performance is slightly better compared to being token-consistent (same $\mathbf{A}$ sample for every token). 
This is surprising as one would expect having more independent noise elements in the stochastic layer would produce a stronger noise effect.
%This is \emph{surprising} as one would expect sampling different noise matrices for different tokens would bring stronger noise.
The choice of noise strength and inference mode will depend on the final use-case. In the following subsections, we analyse the involved trade-offs to guide this decision.

\input{tables/paper/classification}

Next, we insert the stochastic layers into the state-of-the-art Swin transformers~\cite{liu2021swin} to verify that the proposed method is not tailored for a specific ViT architecture. More precisely, we fine tune the Swin-T model, which has a parameter count and computational complexity similar to DeiT-S, following the same fine-tuning procedure as previously described. From Table~\ref{table:imnet_1k_accuracy_swin}, we observe that our stochastic layers behave well across different transformer architectures.

The aim of this paper is to show the benefits of integrating proposed stochastic layers into vision transformers, with just fine-tuning and without architectural modifications. Nevertheless, we want to make sure that the training convergence is not just because of very good initial weights. Therefore, we conduct an experiment where we train from scratch on a 100 randomly chosen classes of the ImageNet-1k dataset. The models are trained followings the settings from~\cite{Touvron21aDeIT}. In Table~\ref{table:imnet_100_scratch_accuracy}, we see that there is no problem learning with our stochastic layers. 

%%%%%%%%%%%%%%%%%%%%%%%%%%%%%%%%%%%%%%%%%%%%%%%%%%%%%%%%%%%%%%%%%%%%%%%%%%%%%%%%%%%%%%%%%%
\subsection{Effects on Confidence Calibration}
\label{ssec:confidence_calibration}

\subsubsection{Problem Definition.}
In practice, any inference algorithm is only useful for a decision-making system, if it can also give an indication about the confidence of its predictions. Intuitively, if a well-calibrated classifier outputs a prediction with a confidence of $p$ (e.g. softmax score), we expect the classifier to be correct with probability $p$. In the context of supervised multi-class classification, random variables $X \in \mathcal{X}$ and $Y \in \mathcal{Y}$ represent the input and labels respectively. Then a classifier $h(X)=(\hat{Y},\hat{P})$, predicting a label $\hat{Y}$ and an associated confidence $\hat{P}$, would be perfectly calibrated if $\mathbb{P}(\hat{Y}=Y | \hat{P} = p) = p \quad \forall p \in \left[0,1\right]$. In particular, we are interested in quantifying the quality of calibration. To this end, we analyse the expected calibration error (ECE), $\mathbb{E}_{\hat{P}}\left[\mathbb{P}(\hat{Y}=Y | \hat{P} = p) - p \right]$. It measures the expected disparity between prediction confidence and accuracy, and would be 0 in the case of perfect calibration. To estimate this value using finite samples,~\cite{Naeini2015ObtainingWC} sorts network predictions $\hat{P}$ into $M$ equidistant bins. Then the ECE is approximated by computing the disparity between accuracy and confidence for samples from each bin as, $ \text{ECE} = \sum_{m=1}^{M}\frac{|B_m|}{n} \left|\text{acc}(B_m) - \text{conf}(B_m)\right|$, where $B_m$ denotes the indices of all samples in the interval $\left(\frac{m-1}{M},\frac{m}{M}\right]$, $\text{acc}(B_m) = \frac{1}{|B_m|}\sum_{i \in B_m} \mathbf{1}(\hat{y}_i=y_i)$ calculates the accuracy on that interval and $\text{conf}(B_m) = \frac{1}{|B_m|}\sum_{i \in B_m} \hat{p}_i$ calculates the confidence on that interval. For our experiments, we compare the ECE with $M=15$ bins ~\cite{guo2017calibration} on the ImageNet-1k validation set.

\input{tables/paper/calibration}

\subsubsection{Experiments.} 
We evaluate the calibration of vision transformers, trained with the procedure described in Section~\ref{sec:experiments_stochastic_layers}, by comparing the ECE metric of token-consistent stochastic layers to the respected baselines.
One very effective method for improving network calibration is the temperature scaling, where the optimal softmax temperature is usually chosen on the validation set~\cite{guo2017calibration}. This method can be used on top of any classification model with a softmax output, and was recently shown to work well with transformers~\cite{Minderer2021RevisitingTC}.
Monte Carlo dropout~\cite{Gal2016DropoutAA} is one of the most popular methods for improving calibration, therefore one baseline is placing dropout layers instead of token-consistent stochastic layers. To create a fair counterpart to our model with a noise level $\Delta$, we chose the dropout probability such that it has the same mean and variance. %as $\mathcal{P}$ from~\eqref{eq:noise_injection}.
Recently proposed masksembles~\cite{Durasov21Masksembles} are the SotA method for improving calibration in CNNs, so another baseline is placing masksemble layers instead of token-consistent stochastic layers.
Finally, one more baseline is the non-token-consistent stochastic layers, where the noise matrix $\mathbf{A}$ from~\eqref{eq:noise_injection} is sampled differently for different tokens.
%The strongest baseline would be to use model ensembles~\cite{lakshminarayanan2017ensambleUncertanity}. However, the computations are very demanding on big datasets, and an ensamble can be made out of any best solution to furthure improve it, so we do not include it as a baseline.

The results of the experiments are summarized in Table~\ref{table:calibration_results}.
We observe that using temperature scaling in any setup improves the calibration. Also, we see that our token-consistent stochastic layers bring better calibration than other baselines, without severe impacts on image recognition. This effect is stronger as the strength of the noise increases, as well as when using Monte Carlo sampling.

%%%%%%
% TRF + Temp + Ours
% Other TRF + Temp
% Some CNN + Temp
% Some CNN + some technique (no TEMP) roundedn to 2 decimals

%%%%
%%%%%%%%%%%%%%%%%%%%%%%%%%%%%%%%%%%%%%%%%%%%%%%%%%%%%%%%%%%%%%%%%%%%%%%%%%%%%%%%%%%%%%%%%%
\subsection{Effects on Adversarial Robustness}
\label{sec:adversarial_robustness}

\subsubsection{Problem definition.}
Besides accuracy and calibration, another important aspect of the network is its behaviour with respect to adversarial examples~\cite{Szegedy2014IntriguingPO,Goodfellow2015ExplainingAH}. 
In general, adversarial attacks exploit the differentiability of the network $f(x)$ and its prediction loss $\mathcal{L}(f(x),l)$, with respect to the input image $x$, where $l$ is the label. Given the true label $l_t$, the attack aims to slightly modify the input $x$ to maximize the prediction loss for label $l_t$. To craft stronger adversarial samples, the fast gradient sign method (FGSM)~\cite{Goodfellow2015ExplainingAH} is repeated multiple times with step size $\alpha$, followed by a projection to an $\epsilon$ hypercube around the initial sample $x$, $\hat{x}^{k} = \Pi_{\epsilon} \left[x^{k-1} + \alpha \, \text{sgn}(\nabla_x \mathcal{L}(f(\hat{x}^{k-1}), l))\right]$. This is known as the projected gradient descent (PGD) attack~\cite{Madry2018TowardsDL}.

\input{tables/paper/adversarial_robustness}

\subsubsection{Evaluation protocol.} 
In order to evaluate adversarial robustness, we craft adversarial examples for each image in the ImageNet-1k validation set, and test the model's accuracy on them. We do this by using the PGD attack with $10$ iterations (PGD-10). When crafting each adversarial example, we initialize the attack's starting point randomly inside the $\epsilon-$hypercube. We restart this procedure $5$ times to find the strongest attack and test for two different $\epsilon$ constraints $\epsilon= \{ \frac{2}{255}, \frac{4}{255}\}$, following the standard benchmark of~\cite{Wong2020FastIB,Shafahi2019AdversarialTF}. The attack step size $\alpha$ is $1$ in both cases.
When dealing with noise inside the network, the expectation-over-transformation (EoT) attack is usually more effective~\cite{Athalye2018ObfuscatedGG}, since it exploits multiple noise samples to avoid gradient obfuscation and to find a common weakness. We also use it during evaluation with 5 expectation samples.
%However, it comes with a very high cost since for every step of the attack one needs to compute multiple forward and backward passes for the same input, in order to estimate the gradient in a Monte Carlo fashion \textcolor{red}{Danda: not clear why we raise this issue?}.  

% \textcolor{red}{Furthermore, it has been shown that many of the stochasticity-based defense strategies reduce the effect of adversarial attacks only because they obfuscate the gradients~\cite{Athalye2018ObfuscatedGG}: Danda: This seems to be a difficult start to defend!}. Since most of the popular attacks use the gradients, it is argued that these defenses would fail for other stronger and more tailored attacks. On one hand, there are scenarios where this is problematic. On the other hand, there are plausible scenarios where \textcolor{red}{the attacker would be unaware of the noise}, thus making a stochasticity-based defense successful. For example, our stochastic layers can be easily introduced to most of the popular transformer models, without modifying the acrhitecture. The attackers could obtain the model weights, but they could still be oblivious to the presence of noise. They would think that it is the original architecture and use it along with the obtained weights to perform an optimization-based attack. \textcolor{red}{Danda: This paragraph is only speaking mind, without clear agenda. In current form, we are better off by removing the whole paragraph..}
% \PR{shall we write a small discussion section after the results?}

\subsubsection{Experiments.} 
Adversarial training is a very effective remedy for adversarial vulnerability ~\cite{Madry2018TowardsDL}, therefore we use it in our experiments. More specifically, we use the efficient adversarial training described in~\cite{Wong2020FastIB}, which fits in our standard training protocol without much overhead. During this robust fine-tuning, regular hyperparameters remain the same as in Section~\ref{sec:experiments_stochastic_layers}, except for using $20$ epochs now, because the optimization is more computationally demanding. Following~\cite{Wong2020FastIB}, the step sizes during adversarial training are $\alpha= \{\frac{2.5}{255}, \frac{5}{255} \}$ for $\epsilon= \{ \frac{2}{255}, \frac{4}{255}\}$, respectively. 
One baseline is applying adversarial training during fine-tuning to the DeiT-S starting point described in Section~\ref{sec:experiments_stochastic_layers}. 
Furthermore, Parametric Noise Injection (PNI)~\cite{Rakin2019ParametricNI} is a SotA adversarial defense when it comes to stochasticity based defenses. Therefore, another baseline is replacing our stochastic layers for PNI layers in the vision transformer.
In Table~\ref{tab:adv_robustness_sota} we see that our token-consistent stochastic layers improves adversarial robustness beyond adversarial training. We also see that a better robustness is achieved compared to using the PNI baseline. Furthermore, introducing our stochastic layers brings small adversarial robustness, even without adversarial training.

Next, in Table~\ref{tab:adv_robustness_thorough} we see an analysis of the effects of token-consistent stochastic layers on robustness. 
The network is more robust with stronger noise and when networks are sampled in a Monte Carlo fashion. Also, the clean accuracy becomes higher with sampling. This is particularly interesting, since it has been shown that there is a trade-off between adversarial robustness and clean accuracy~\cite{Zhang2019TheoreticallyPT}. 
Moreover, under the EoT attack, our stochastic layers perform better than the SotA PNI~\cite{Rakin2019ParametricNI}. Unfortunately though, neither method performs significantly better than the deterministic baseline. This may be attributed to the possibility of being able to estimate a reliable expectation of the gradients by means of Monte Carlo sampling. In fact, alternative methods could be designed to hinder the expectation estimation required for the EoT attack. More importantly, the EoT attack is significantly more computationally demanding than the PGD attack, making the attack itself more difficult to perform.     

\subsection{Effects on Privacy in Collaborative Settings}
\label{ssec:private_features}

% We first analyse to which extent vision transformers preserve the privacy of features. To this end, we train a decoder to reconstruct the input from features of different transformer blocks, and show the impact of our stochastic layers on reconstruction quality, 
% %sample complexity of the reconstruction problem, 
% and quantify the trade-off between accuracy and noise level.
% Then, we further demonstrate that the stochastic network still retains its generalization ability and flexibility to transfer to other datasets, and is therefore a directly applicable for federated learning, while ensuring a higher level of privacy. 
% %\textcolor{red}{ This subsection looks poor due to the lack  or too much of fragmentation.}

\subsubsection{Problem definition.}
\label{ssec:input_reconstruction}
We consider a practical collaborative learning and inference setup $\hat{y}=\Phi_2(\Phi_1(x))=\Phi(x)$, where $\Phi_1$ is the client network executed on trusted devices and $\Phi_2$ is a task network executed on an untrusted server using the clients' activation $z=\Phi_1(x)$~\cite{Singh2021DISCODA,Roy2019MitigatingIL,Li2021DepObfuscation}.
%To simulate this setup, we split the model into two parts.
The client network $\Phi_1$ is the first part of the network $f$ which produces the feature map $z=\mathcal{\Tilde{Z}}^{FC_1}_{i}$ from~\eqref{eq:noise_injection} of the $i$-th block. The client sends $z$ to the server $\Phi_2$, which is the second part of the network $\Phi$.
For collaborative inference, the server then predicts $\hat{y}=\Phi_2(z)$.

For collaborative learning, the client sends labelled $(z, y)$ pairs to the server. In general, the server trains the network $\Phi_2$, and broadcasts the gradient to update $\Phi_1$~\cite{vepakomma2018split}. We use a slightly different setting, where $\Phi_1$ is a strong general feature extractor (e.g. pretrained on ImageNet), whose features allow for effective transfer to the desired tasks~\cite{Kornblith2019DoBI}. In this scenario, the $\Phi_1$ does not update, while only $\Phi_2$ is trained on the server using the data $(z, y)$ from all participants. 

Under our threat model, the untrusted server attempts to obtain sensitive information by recovering the input image $x$ (e.g. a face or medical image). 
The input $x$ is recovered from the client activation $z=\mathcal{\Tilde{Z}}^{FC_1}_{i}$, by using a reconstruction network $\hat{x} = r(z)$. 
The practical validity of this threat model is in the scenarios where a small number of training pairs $(x, z)$ is obtained through a malicious or colluding client, who is participating in the collaborative setting~\cite{Singh2021DISCODA}. Another scenario is when pairs $(x, z)$ from a similar distribution are publicly available or intercepted. 
Therefore, it is very important for the shared feature map $z$ to be private.
%\textcolor{red}{DANDA: maybe the context of the paper and this setup is little detached. Not sure though..}

% \noindent\textbf{Threat Model.} The adversary intercepts feature maps $\mathcal{\Tilde{Z}}^{FC_1}_{i}$, while they are being sent to the server. Similarly, an adversary could potentially obtain a small amount of input images corresponding to the sent feature maps and use them to learn a model which reconstructs the input image. When a new feature map $\mathcal{\Tilde{Z}}^{FC_1}_{i}$ is sent to the server, the adversary is now able to reconstruct the corresponding input image and thus violate the users privacy.

\input{figures/paper/collaborative_inference}
%\input{figures/paper/privacy_reconstruction_uniformn}
%\input{figures/paper/privacy_reconstruction_uniform_everywhere}

\subsubsection{Experiments.}
In order to evaluate the privacy of shared features in the collaborative inference, we train a decoder to reconstruct the input image from the feature map $z=\mathcal{\Tilde{Z}}^{FC_1}_{i}$, from different blocks $i$ of the models trained in Section~\ref{sec:experiments_stochastic_layers}.
A good reconstruction quality indicates a weak protection of sensitive information, and thus low privacy.
The decoder from~\cite{isola2018imagetoimage} is used, after adjusting it to the spatial size of $\mathcal{\Tilde{Z}}^{FC_1}_{i}$.
We use the $L_1$ reconstruction loss to train the decoder on a small subset of ImageNet-1k, containing $20$ randomly chosen classes. 
We evaluate the reconstruction quality by measuring the $L_1$, $L_2$, peak signal to noise ratio (PSNR) and structural similarity index measure (SSIM) metrics.
In Figures~\ref{fig:privacy_L1}~-~\ref{fig:privacy_SSIM} we see that it is harder to reconstruct the input from feature maps $\mathcal{\Tilde{Z}}^{FC_1}_{i}$ when using our stochastic layers. The gap to the deterministic network increases with stronger levels of noise. 
Next, in Figures~\ref{fig:privacy_dropout}~-~\ref{fig:privacy_uniform_everywhere} we see that our token-consistent stochastic layers preserve more privacy compared to regular dropout or the non-token-consistent version.
These experimental results demonstrate the privacy benefits of our token-consistent stochastic layers for collaborative inference, on the basis of poor reconstruction quality. 
The additional comparisons with the related works in collaborative inference is beyond the scope of this paper, as they use tailored architectures, losses, and learning setups designed for the same~\cite{Singh2021DISCODA,Roy2019MitigatingIL,Li2021DepObfuscation}.

 In the case of collaborative learning, we train only $\Phi_2$ on a new task of CIFAR10/100~\cite{Krizhevsky2009LearningML} image classification. We model $\Phi_2$ by two fully connected and one softmax layers, to address the task at hand.  In Figure~\ref{fig:privacy_split_learning} we visualize the trade-off between the accuracy on the new tasks and the input reconstruction (feature privacy). We observe that using stochastic layers offers better privacy in this setting at a cost of task performance. With this knowledge one can choose a trade-off point, in accordance with the final task and the use-case.

\input{figures/paper/privacy_split_learning}

%%%%%%%%%%%%%%%%%%%%%%%%%%%%%%%%%%%%%%%%%%%%%%%%%%%%%%%%%%%%%%%%%%%%%%%%%%%%%%%%%%%%%%%%%%
\subsection{Effects on Transfer Learning}
\label{sec:transfer_learning}

\input{tables/paper/transfer_learning}

To measure the impact of our stochastic layers on general image recognition performance, we compare the transferability of our features on a standard benchmark suite following the setup of~\cite{salman2020adveTransfer}. 
We use the AdamW optimizer~\cite{Loshchilov2017AdamW} with an inital learning rate of $10^{-5}$ and weight decay of $5\times10^{-2}$. The transfer learning lasts for $150$ epochs and the learning rate gets decayed by a factor of $10$ every $50$ epochs.
Our findings are summarized in Table~\ref{table:transfer_learning}. Despite the tendency of big transformer architectures to quickly overfit, all compared method obtain competitive results. We observe that our token-consistent stochastic layers are able to retain their performance and sometimes even exceed the deterministic baseline. The transferability is retained across different noise levels. 
%We therefore conclude that the transferability is not negatively impacted by introducing the stochasticity.\textcolor{red}{DANDA: we end the paragraph with more positive note.}

% One step closer to binarization/compression (with results)

%% file: figures/paper/classification.tex
\begin{figure}[t!]
    \label{fig:main_teaser}
    \centering
    \captionsetup{font=small}
    \centering
    \includegraphics[width=0.65\columnwidth]{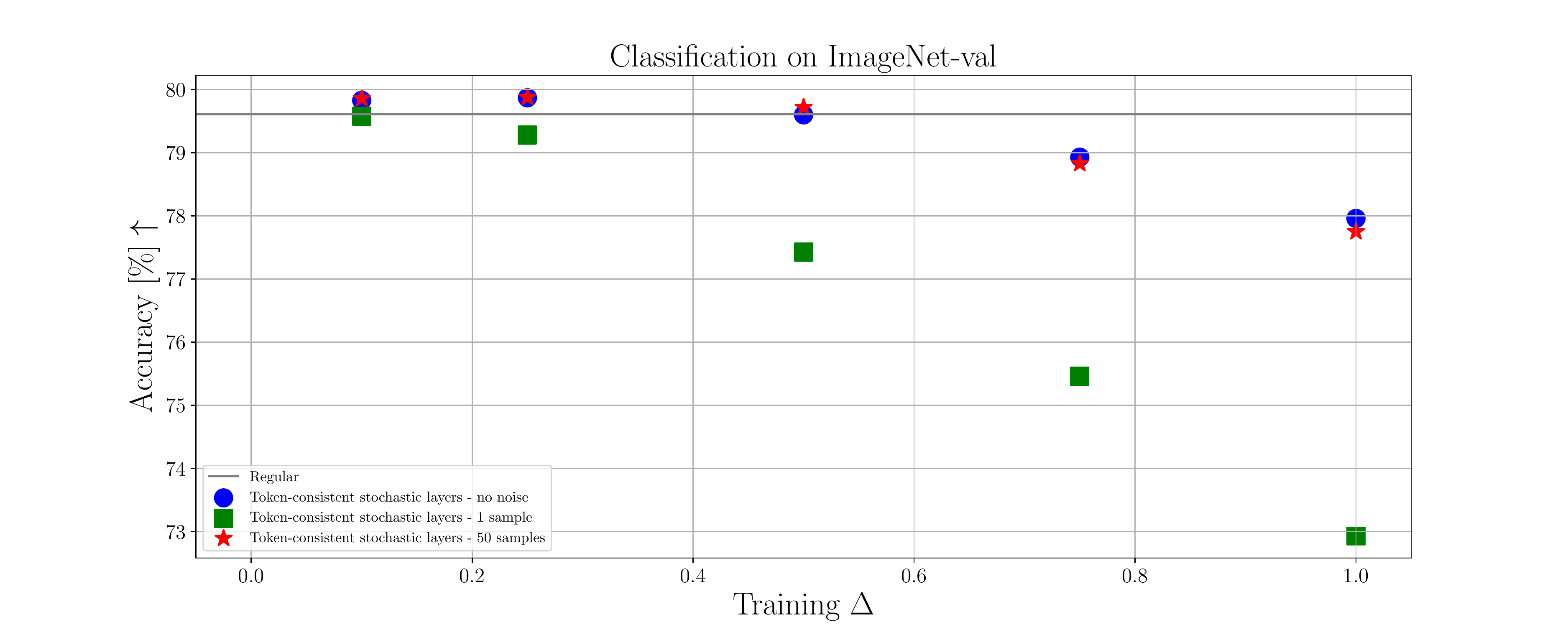}
    \vspace{-4pt}
    \caption{\textbf{ImageNet classification.} Token-consistent stochastic layers bring no significant drop in accuracy, even with strong levels of noise. Performance is better when $a^{j}$ is fixed to the mean, or when using Monte Carlo sampling during inference. }
    \label{fig:classification_accuracy}
    % \caption{\textbf{An overview of the effects of proposed stochastic layers in vision transformers.} The channel $j$ of every token inside the transformer's MLP block is multiplied with the same random variable $a^{j} \sim \mathcal{U}(1-\Delta,1+\Delta)$ from~\eqref{eq:noise_injection}. In~\ref{fig:main_teaser_accuracy} we see that there is no significant drop in accuracy, even when using strong stochasticity during inference. 
    % Next, in~\ref{fig:main_teaser_calibration} we see that adding stochastic layers can improve network calibration, meaning that the predictive probability better depicts the uncertainty.
    % In~\ref{fig:main_teaser_adv_robustness} we can see that adding stochastic layers can improve robustness against adversarial attacks.  
    % Finally, in~\ref{fig:main_teaser_privacy} we see that adding stochastic layers can help with privacy preservation by reducing reconstruction quality of feature maps sent over unsecure channels.}
\end{figure}

%% file: tables/paper/classification.tex
\begin{table}[t!]
\centering
\tableFont
\captionsetup{font=small}
\caption{\textbf{ImageNet classification.} Performance of token-consistent stochastic layers.}
\vspace{-5pt}
%%%%%%%%%%%%%%%%%%%%%%%%%%%%%%%%%%%%%%%%%%%%%%%%%%%%%%%%%%%%%%%%%%%%%%%%%%%%%%
\begin{subtable}{1\columnwidth}
%\textbf{ImageNet-100 classification from scratch}.
\caption{
Using the same sample of $\mathbf{A}$ from~\eqref{eq:noise_injection} for different tokens (token-consistent) is a stronger form of noise compared to sampling different $\mathbf{A}$ for different tokens.}
%Using different samples of the noise matrix $\mathbf{A}$ for different tokens is a weaker form of noise compared to using the same $\mathbf{A}$ sample for every token (token-consistent).}
\centering
\begin{tabular}{|l|l|c|c|c|}
\hline
\multicolumn{2}{|c|}{} & \multicolumn{3}{c|}{Accuracy$\uparrow$}  \\ \cline{3-5}
\multicolumn{2}{|c|}{} & Noise off & $N=1$ & $N=50$         \\ \hline \hline

\multicolumn{2}{|l|}{DeiT-S}  & \multicolumn{3}{c|}{$79.61\%$}  \\ \hline \hline

\multirow{3}{*}{\makecell{$\Delta=0.5$}}  
 & DeiT-S + dropout~\cite{srivastava2014dropout} & $79.60\%$ & $78.54\%$ & $79.79\%$ \\ \cline{2-5}
 & DeiT-S + not-token-consistent uniform & $79.63\%$ & $78.59\%$ & $79.81\%$ \\ \cline{2-5}
 & DeiT-S + token-consistent uniform & $79.60\%$ & $77.43\%$ & $79.72\%$ \\ \hline \hline
 
\multirow{3}{*}{\makecell{$\Delta=1.0$}}   
 & DeiT-S + dropout~\cite{srivastava2014dropout} & $77.93\%$ & $76.18\%$ & $78.74\%$ \\ \cline{2-5}
 & DeiT-S + not-token-consistent uniform & $77.99\%$ & $75.85\%$ & $78.63\%$ \\ \cline{2-5}
 & DeiT-S + token-consistent uniform  & $77.96\%$ & $72.93\%$ & $77.75\%$ \\ \hline

\end{tabular}

\label{table:imnet_1k_accuracy_thorough}
\end{subtable}
%%%%%%%%%%%%%%%%%%%%%%%%%%%%%%%%%%%%%%%%%%%%%%%%%%%%%%%%%%%%%%%%%%%%%%%%%%%%%%

%%%%%%%%%%%%%%%%%%%%%%%%%%%%%%%%%%%%%%%%%%%%%%%%%%%%%%%%%%%%%%%%%%%%%%%%%%%%%%
\begin{subtable}{1\columnwidth}
%\textbf{Different transformer architectures.}
\caption{Token-consistent stochastic layers also behave well on different architectures.}
\centering

\begin{tabular}{|l|l|c|c|c|}
\hline
\multicolumn{2}{|c|}{} & \multicolumn{3}{c|}{Accuracy$\uparrow$}  \\ \cline{3-5}
\multicolumn{2}{|c|}{} & Noise off & $N=1$ & $N=50$         \\ \hline \hline
\multirow{4}{*}{DeiT-S~\cite{Touvron21aDeIT}} 
 & no noise & \multicolumn{3}{c|}{$79.61\%$} \\ \cline{2-5}
 & + $\Delta=0.1$ & $79.83\%$ & $79.58\%$ & $79.86\%$ \\ \cline{2-5}
 & + $\Delta=0.5$ & $79.60\%$ & $77.43\%$ & $79.72\%$ \\ \cline{2-5}
 & + $\Delta=1.0$ & $77.96\%$ & $72.93\%$ & $77.75\%$ \\ \hline \hline 
\multirow{4}{*}{Swin-T~\cite{liu2021swin}} 
 & no noise & \multicolumn{3}{c|}{$81.03\%$} \\ \cline{2-5}
 & + $\Delta=0.1$ & $81.02\%$ & $81.02\%$ & $81.09\%$ \\ \cline{2-5}
 & + $\Delta=0.5$ & $81.05\%$ & $79.40\%$ & $81.07\%$ \\ \cline{2-5}
 & + $\Delta=1.0$ & $80.20\%$ & $75.95\%$ & $80.08\%$ \\ \hline 
 
\end{tabular}
\label{table:imnet_1k_accuracy_swin}
\end{subtable}
%%%%%%%%%%%%%%%%%%%%%%%%%%%%%%%%%%%%%%%%%%%%%%%%%%%%%%%%%%%%%%%%%%%%%%%%%%%%%%

%%%%%%%%%%%%%%%%%%%%%%%%%%%%%%%%%%%%%%%%%%%%%%%%%%%%%%%%%%%%%%%%%%%%%%%%%%%%%%
\begin{subtable}{1\columnwidth}
%\textbf{ImageNet-100 classification from scratch}.
\caption{The networks with token-consistent stochastic layers are also successfully trained from scratch on $100$ randomly chosen classes of the ImageNet-1k dataset.}
\centering

\begin{tabular}{|l|l|c|c|c|}
\hline
\multicolumn{2}{|c|}{} & \multicolumn{3}{c|}{Accuracy$\uparrow$}  \\ \cline{3-5}
\multicolumn{2}{|c|}{} & Noise off & $N=1$ & $N=50$         \\ \hline \hline

\multicolumn{2}{|l|}{DeiT-S}  & $89.18\%$ & $89.18\%$ & $89.18\%$  \\ \hline 

\multicolumn{2}{|l|}{DeiT-S + $\Delta=0.1$}                       & $89.40\%$ & $89.30\%$ & $89.28\%$  \\ \hline 

\multicolumn{2}{|l|}{DeiT-S + $\Delta=0.5$}                       & $89.48\%$ & $88.58\%$ & $89.48\%$  \\ \hline 

\multicolumn{2}{|l|}{DeiT-S + $\Delta=1.0$}                       & $89.06\%$ & $88.00\%$ & $89.24\%$  \\ \hline

\end{tabular}
%}
\label{table:imnet_100_scratch_accuracy}
\end{subtable}
%%%%%%%%%%%%%%%%%%%%%%%%%%%%%%%%%%%%%%%%%%%%%%%%%%%%%%%%%%%%%%%%%%%%%%%%%%%%%%

\end{table}

%% file: tables/paper/calibration.tex
\begin{table}[t!]
\centering
\tableFont
\captionsetup{font=small}
\caption{\textbf{Confidence calibration.} Token-consistent stochastic layers achieve better calibration compared to several respected baselines and non-token-consistent uniform noise. The calibration is better for stronger noise, and with Monte Carlo sampling.}
\centering
\begin{tabular}{|l|l|c|c|c|}
\hline
\multicolumn{2}{|c|}{} & \multicolumn{3}{c|}{ECE$\downarrow$ (Accuracy$\uparrow$)}  \\ \hline \hline

% Accuracy $79.61\%$
\multicolumn{2}{|l|}{DeiT-S}  & \multicolumn{3}{c|}{$0.088 (79.8\%)$}  \\ \hline 

% Accuracy $79.61\%$
\multicolumn{2}{|l|}{DeiT-S + temp.~\cite{guo2017calibration}}  & \multicolumn{3}{c|}{$0.026 (79.8\%)$}  \\ \hline 

% % % Accuracy $\%$
% \multicolumn{2}{|l|}{DeiT-S + masksembles~\cite{Durasov21Masksembles}}  & \multicolumn{3}{c|}{$0.121 (76.7\%)$}  \\ \hline

% Accuracy $76.73\%$
\multicolumn{2}{|l|}{DeiT-S + masksembles~\cite{Durasov21Masksembles}  + temp.~\cite{guo2017calibration}}  & \multicolumn{3}{c|}{$0.017 (76.7\%)$}  \\ \hline \hline

% % Accuracy $x\%$
% \multicolumn{2}{|l|}{x}  & \multicolumn{3}{c|}{$x$}  \\ \hline 

\multicolumn{2}{|l|}{DeiT-S + stochastic layers + temp.~\cite{guo2017calibration} } & Noise off  & $N=1$ & $N=50$         \\ \hline 

\multirow{3}{*}{\makecell{$\Delta=0.5$}} 
 & dropout~\cite{Gal2016DropoutAA}& $0.022 (79.6\%)$ & $0.020 (78.7\%)$ & $0.019 (79.7\%)$ \\ \cline{2-5}
 & not-token-consistent uniform& $0.023 (79.6\%)$ & $0.022 (78.6\%)$ & $0.019 (79.8\%)$  \\ \cline{2-5}
 & token-consistent uniform & $0.023 (79.6\%)$ & $0.020 (77.5\%)$ & $0.017 (79.6\%)$ \\ \hline
 
\multirow{3}{*}{\makecell{$\Delta=1.0$}} 
 & dropout~\cite{Gal2016DropoutAA}& $0.019 (77.9\%)$ & $0.019 (76.1\%)$ & $0.014 (78.7\%)$ \\ \cline{2-5}
 & not-token-consistent uniform& $0.019 (80.0\%)$ & $0.016 (76.0\%)$ & $0.014 (78.6\%)$ \\ \cline{2-5} 
 & token-consistent uniform& $0.016 (78.0\%)$ & $0.012 (73.1\%)$ & $\mathbf{0.010} (77.6\%)$ \\ \hline

\end{tabular}

\label{table:calibration_results}
\end{table}

%% file: tables/paper/adversarial_robustness.tex
\begin{table}[t!]
\centering
\tableFont
\captionsetup{font=small}
\caption{\textbf{Adversarial robustness}. 
In Table~\ref{tab:adv_robustness_sota}, we see that our stochastic layers are more robust compared to the evaluated baselines.
Table~\ref{tab:adv_robustness_thorough} shows that using Monte Carlo sampling with our stochastic layers improves adversarial robustness and clean accuracy, under adversarial training. This effect is stronger with stronger noise.}

%%%%%%%%%%%%%%%%%%%%%%%%%%%%%%%%%%%%%%%%%%%%%%%%%%%%%%%%%%%%%%%%%%%%%%%%%%%%%%
\begin{subtable}{1\columnwidth}
\caption{Overview of the adversarial robustness compared to different baselines.}
\centering
%\resizebox{0.8\columnwidth}{!}{
\begin{tabular}{|l||c|c|c|c|}
\hline
& \multicolumn{4}{c|}{Accuracy $\uparrow$} \\ \cline{2-5} 

& \multicolumn{2}{c||}{\makecell{Adversarial training \\ with $\epsilon=\frac{2}{255}$}}
& \multicolumn{2}{c|}{\makecell{Adversarial training \\ with $\epsilon=\frac{4}{255}$}}
\\ \cline{2-5}

& \makecell{Clean \\ samples}
& {\makecell{PDG$-10$ \\ attack}} 
& \makecell{Clean \\ samples}
& {\makecell{PDG$-10$ \\ attack}}
\\ \hline

% % Dnt mention restart num of PGD restarts, probably 1 restart
% ResNet-50 + adv. tr.~\cite{Shafahi2019AdversarialTF} & $64.45\%$ & $43.52\%$ & $60.21\%$ & $32.77\%$ \\ \hline  

% % 1 PGD restart,  almost same for 10 restarts
% ResNet-50 + adv. tr.~\cite{Wong2020FastIB} & $64.37\%$ & $43.31\%$ & $60.42\%$ & $31.22\%$ \\ \hline 

% 5 PGD restarts
DeiT-S & $\textbf{79.61\%}$ & $0.43\%$ & $79.61\%$ & $0.01\%$ \\ \hline 

% 5 PGD restarts
DeiT-S + ours & $77.75\%$ & $12.36\%$ & $77.75\%$ & $4.76\%$ \\ \hline 

% 5 PGD restarts
DeiT-S + adv. tr.~\cite{Wong2020FastIB} & $71.61\%$ & $42.33\%$ & $65.04\%$ & $27.24\%$ \\ \hline 

% 5 PGD restarts
DeiT-S + PNI~\cite{Rakin2019ParametricNI}  + adv. tr.~\cite{Wong2020FastIB}  & $73.83\%$ & $45.16\%$ & $69.24\%$ & $31.18\%$ \\ \hline 

% 5 PGD restarts
DeiT-S + ours  + adv. tr.~\cite{Wong2020FastIB} & $73.83\%$ & $\textbf{49.32\%}$ & $\textbf{70.47\%}$ & $\textbf{36.38\%}$ \\ \hline 

% Compare to regular adv robustness on Imnet on Resnet50 

% https://arxiv.org/pdf/1907.02610.pdf only ahs ReSNet152 with very good results , a lot probably ahs to do that it is a bigger model

\end{tabular}
%}
\label{tab:adv_robustness_sota}
\end{subtable}
%}
%%%%%%%%%%%%%%%%%%%%%%%%%%%%%%%%%%%%%%%%%%%%%%%%%%%%%%%%%%%%%%%%%%%%%%%%%%%%%%%%%%

%%%%%%%%%%%%%%%%%%%%%%%%%%%%%%%%%%%%%%%%%%%%%%%%%%%%%%%%%%%%%%%%%%%%%%%%%%%%%%
%\resizebox{\columnwidth}{!}{
%\renewcommand\arraystretch{1}
\begin{subtable}{1\columnwidth}
\caption{Analysis of the effects of our stochastic layers on adversarial robustness.}
%\resizebox{\columnwidth}{!}{
\centering
\begin{tabular}{|c|l||c|c|c||c|c|c|}

\hline
\multicolumn{2}{|c||}{} & \multicolumn{6}{c|}{\makecell{Accuracy $\uparrow$}} \\ \cline{3-8}

%\multicolumn{3}{|c|}{} & \multicolumn{3}{|c|}{\makecell{Without adversarial training}} \\ \cline{4-6}
  
\multicolumn{2}{|c||}{} 
 %& Clean
 & \multicolumn{3}{c||}{\makecell{Adversarial training \\ with $\epsilon=\frac{2}{255}$~\cite{Wong2020FastIB} }} 
 & \multicolumn{3}{c|}{\makecell{Adversarial training \\ with $\epsilon=\frac{4}{255}$~\cite{Wong2020FastIB}}}
 \\ \cline{3-8}
 
\multicolumn{2}{|c||}{} 
 & \makecell{Clean \\ samples}
 & {\makecell{PDG$-10$ \\ attack}} 
 & {\makecell{PDG$-10$ \\ EoT attack}} 
 & \makecell{Clean \\ samples}
 & {\makecell{PDG$-10$ \\ attack}}
 & {\makecell{PDG$-10$ \\ EoT attack}} 
 \\ \hline
 
\multicolumn{2}{|c||}{DeiT-S} & $71.61\%$ & $42.33\%$ & $42.33\%$ & $65.04\%$ & $27.24\%$ & $27.24\%$ \\ \hline 

\multirow{2}{*}{\makecell{ DeiT-S + PNI~\cite{Rakin2019ParametricNI}\\ (layerwise)}}
 & $N=1$  & $73.50\%$ & $44.84\%$ & $40.60\%$ & $68.77\%$ & $31.07\%$ & $25.90\%$ \\ \cline{2-8} 
 & $N=50$ & $73.83\%$ & $45.16\%$ & $40.88\%$ & $69.24\%$ & $31.18\%$ & $26.02\%$ \\ \hline
 
\multirow{2}{*}{\makecell{ DeiT-S + PNI~\cite{Rakin2019ParametricNI}\\ (channelwise)}}
 & $N=1$  & $72.88\%$ & $44.47\%$ & $40.96\%$ & $67.40\%$ & $30.22\%$ & $26.11\%$ \\ \cline{2-8} 
 & $N=50$ & $73.47\%$ & $44.89\%$ & $41.37\%$ & $68.30\%$ & $30.37\%$ & $26.48\%$ \\ \hline
 
\multirow{2}{*}{\makecell{ DeiT-S + PNI~\cite{Rakin2019ParametricNI}\\ (elementwise)}}
 & $N=1$  & $72.51\%$ & $44.49\%$ & $41.08\%$ & $67.02\%$ & $29.99\%$ & $26.13\%$ \\ \cline{2-8} 
 & $N=50$ & $73.41\%$ & $44.85\%$ & $41.49\%$ & $68.01\%$ & $30.20\%$ & $26.54\%$ \\ \hline
 
\multirow{2}{*}{\makecell{ DeiT-S + ours\\ ($\Delta=0.1$)}}
 & $N=1$  & $71.79\%$ & $42.90\%$ & $42.44\%$ & $65.23\%$ & $27.88\%$ & $27.39\%$ \\ \cline{2-8} 
 & $N=50$ & $71.84\%$ & $42.91\%$ & $\textbf{42.53\%}$ & $65.29\%$ & $27.86\%$ & $\textbf{27.46\%}$ \\ \hline

\multirow{2}{*}{\makecell{ DeiT-S + ours\\ ($\Delta=0.5$)}}
 & $N=1$  & $71.79\%$ & $44.51\%$ & $40.29\%$ & $66.53\%$ & $30.68\%$ & $25.80\%$ \\ \cline{2-8} 
 & $N=50$ & $73.73\%$ & $46.21\%$ & $41.64\%$ & $68.65\%$ & $31.47\%$ & $26.58\%$ \\ \hline
 
\multirow{2}{*}{\makecell{ DeiT-S + ours\\ ($\Delta=1.0$)}}
 & $N=1$  & $68.62\%$ & $45.37\%$ & $36.75\%$ & $65.40\%$ & $33.48\%$ & $23.30\%$ \\ \cline{2-8}  
 & $N=50$ & $\textbf{73.83\%}$ & $\textbf{49.32\%}$ & $39.58\%$ & $\textbf{70.47\%}$ & $\textbf{36.38\%}$ & $24.92\%$ \\ \hline
 \multicolumn{2}{|c||}{\makecell{Iterations per attack}} & 0 & 50 & 250 & 0 & 50 & 250 \\ \hline
\end{tabular}
%}
\label{tab:adv_robustness_thorough}
\end{subtable}

\label{tab:adv_robustness}
\end{table}

%% file: figures/paper/collaborative_inference.tex
% For CVPR 2 column format we used \begin{subfigure}[b]{0.47\columnwidth}
% and \includegraphics[width=\textwidth]{img_path}
\begin{figure}[t!]
    \centering
    \captionsetup{font=small}
    \begin{subfigure}[b]{0.23\columnwidth}
        \centering
        \includegraphics[width=\columnwidth]{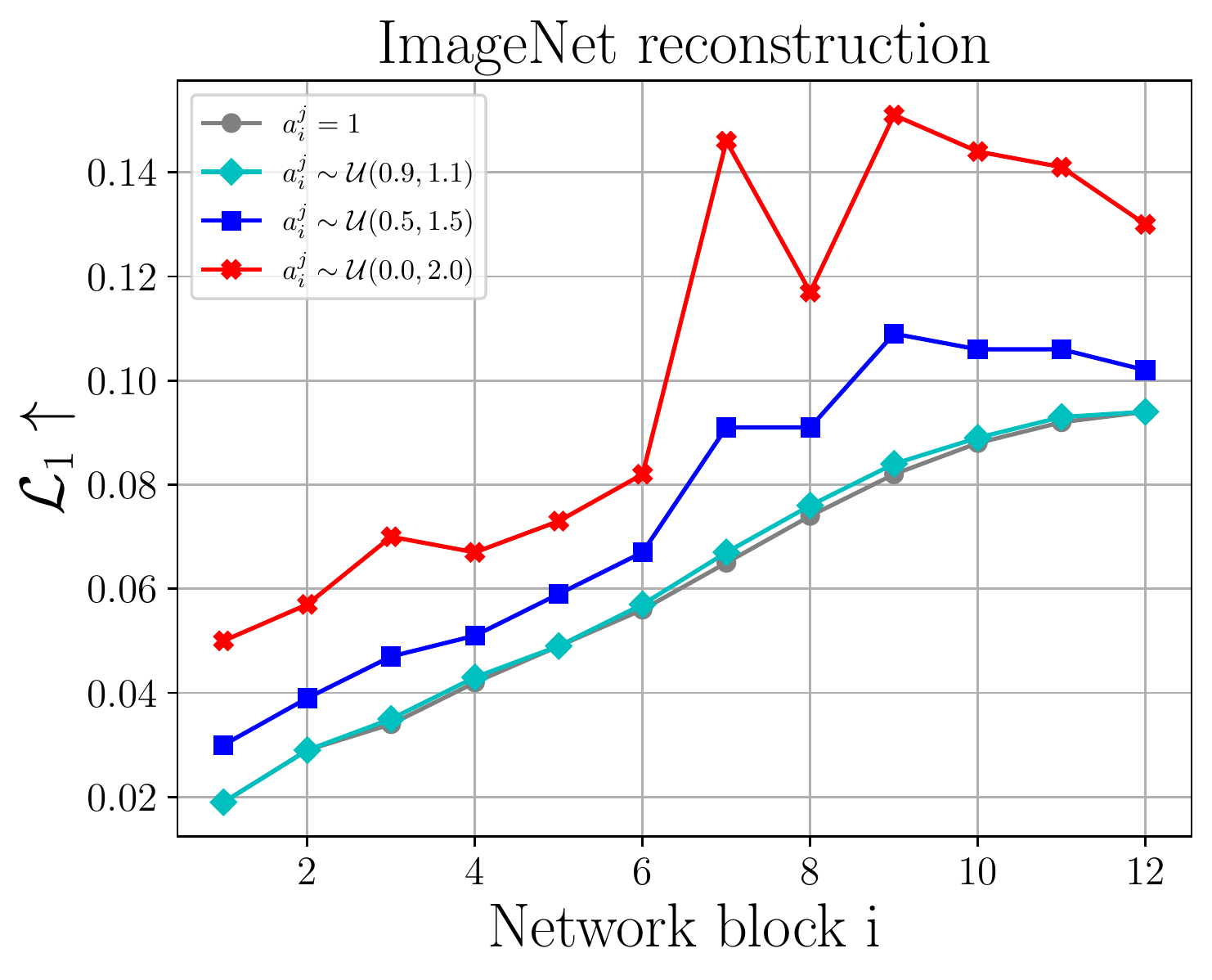}
        \caption{}
        %\caption{$L_1$ reconstruction}
        \label{fig:privacy_L1}
    \end{subfigure}
    %\hfill    
    \begin{subfigure}[b]{0.23\columnwidth}
        \centering
        \includegraphics[width=\columnwidth]{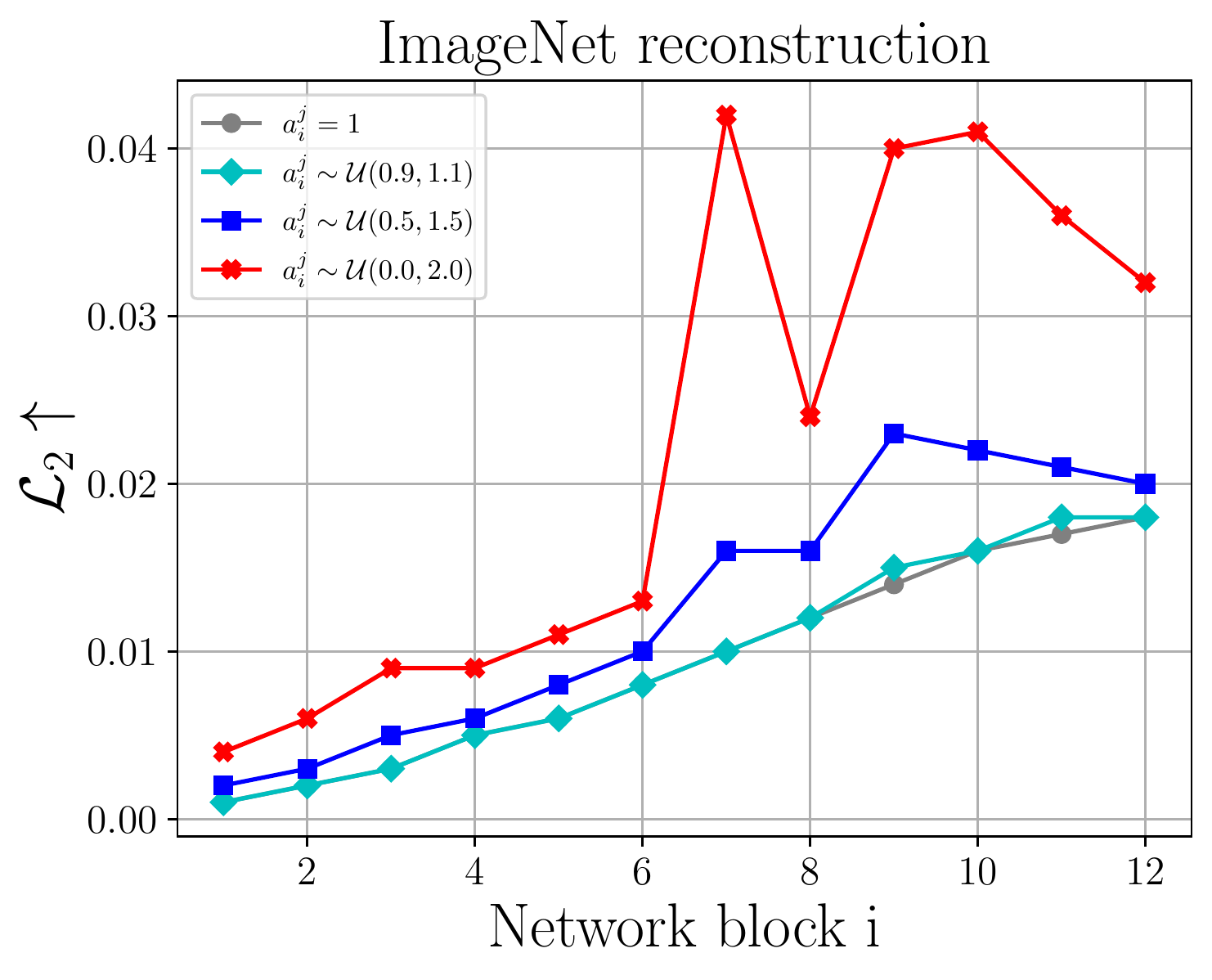}
        \caption{}
        %\caption{$L_2$ reconstruction}
        \label{fig:privacy_L2}
    \end{subfigure}
    %\hfill
    %\vskip\baselineskip
    \begin{subfigure}[b]{0.23\columnwidth}
        \centering
        \includegraphics[width=\columnwidth]{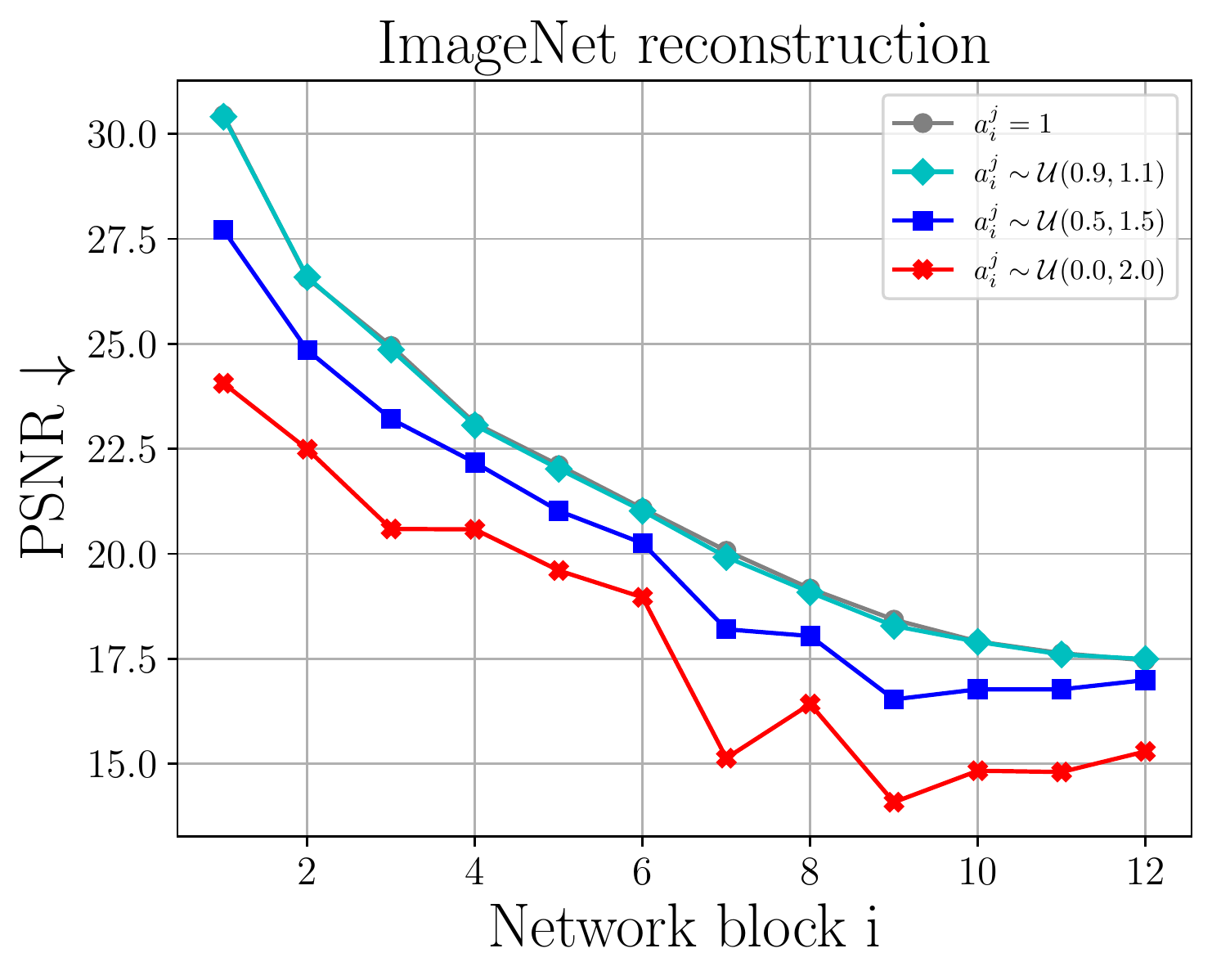}
        \caption{}
        %\caption{PSNR reconstruction}
        \label{fig:privacy_PSNR}
    \end{subfigure}
    %\hfill    
    \begin{subfigure}[b]{0.23\columnwidth}
        \centering
        \includegraphics[width=\columnwidth]{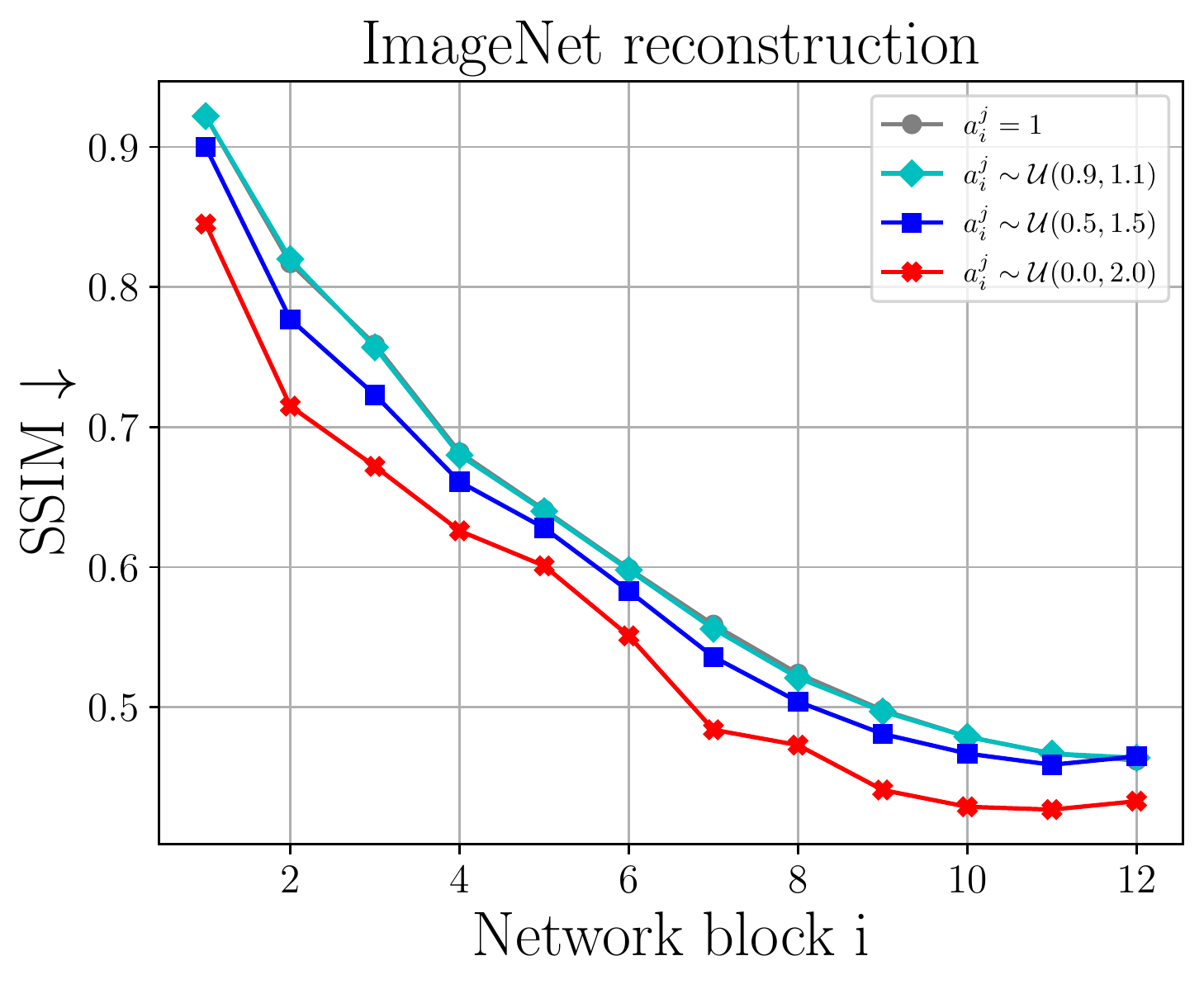}
        \caption{}
        %\caption{SSIM reconstruction}
        \label{fig:privacy_SSIM}
    \end{subfigure}

    \begin{subfigure}[b]{0.23\columnwidth}
        \centering
        \includegraphics[width=\columnwidth]{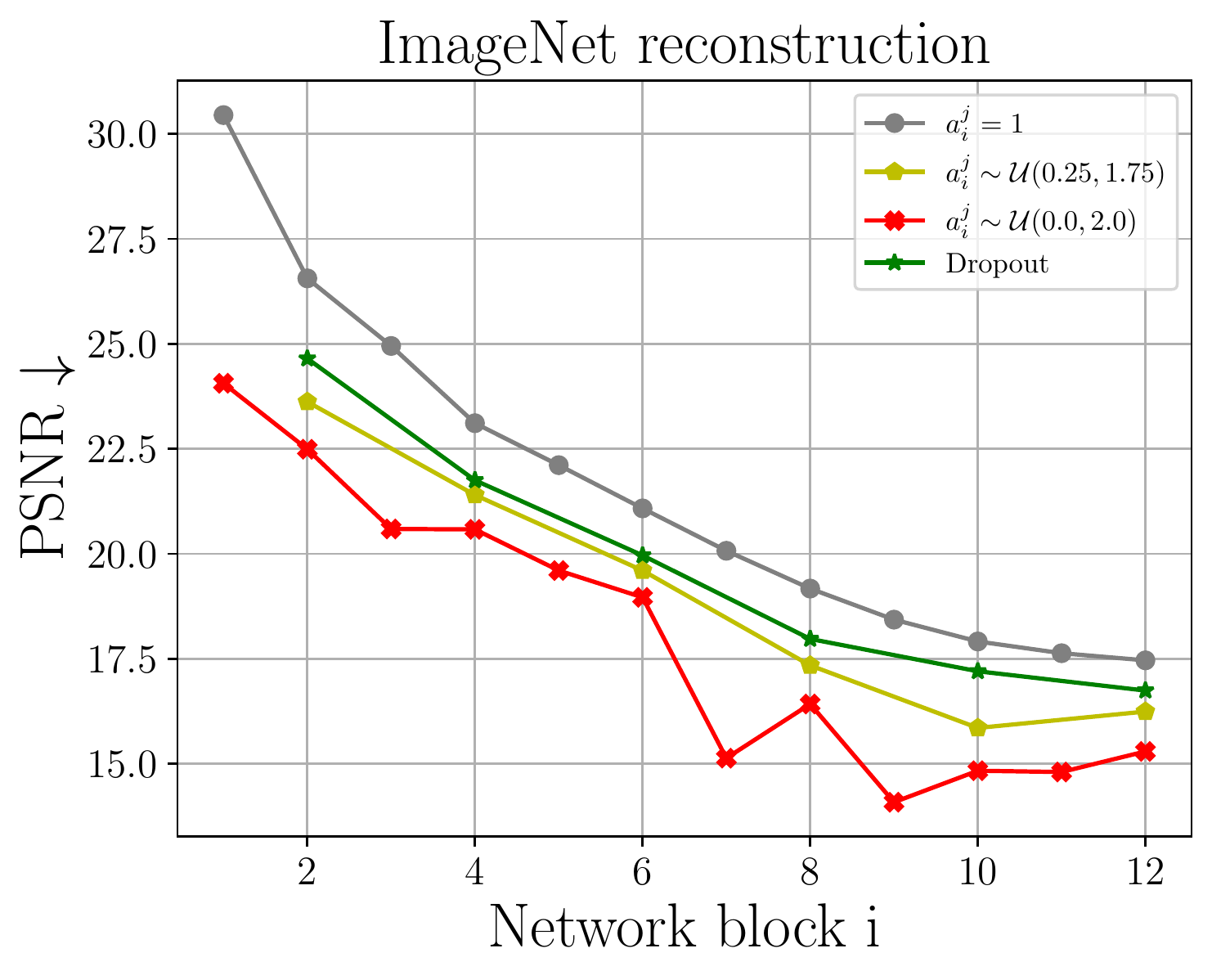}
        \caption{}
        %\caption{Dropout}
        \label{fig:privacy_dropout}
    \end{subfigure}
    %\hfill    
    \begin{subfigure}[b]{0.23\columnwidth}
        \centering
        \includegraphics[width=\columnwidth]{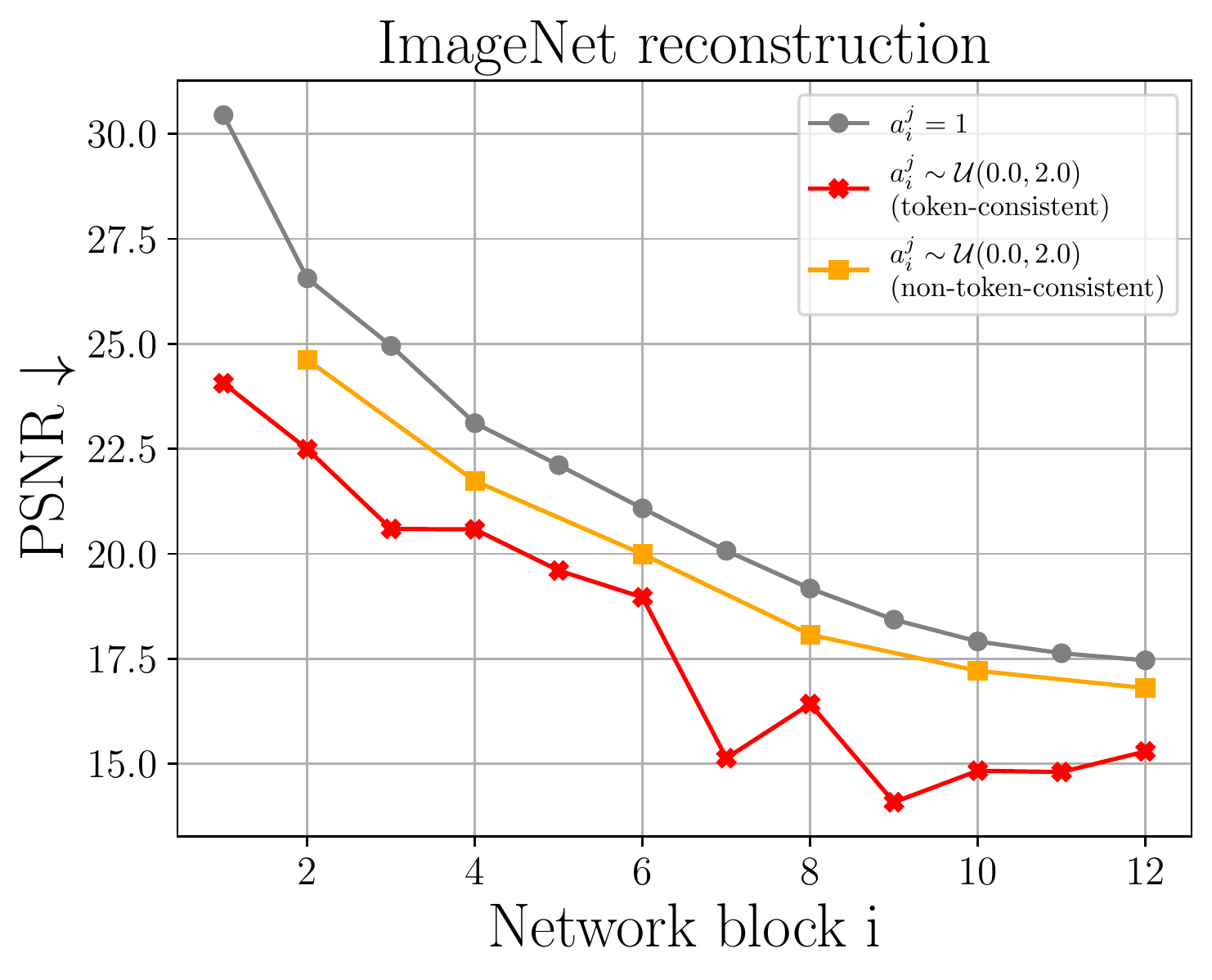}
        \caption{}
        %\caption{Uniform}
        \label{fig:privacy_uniform_everywhere}
    \end{subfigure}
    \vspace{-5pt}
    \caption{{\textbf{Privacy in collaborative inference}. 
     In Figures~\ref{fig:privacy_L1}~-~\ref{fig:privacy_SSIM} we see that stronger noise in our token-consistent stochastic layers makes the features more private.
     In Figures~\ref{fig:privacy_dropout} we see that regular dropout preserves less privacy than our stochastic layers (dropout matches the variance of $\mathcal{U}(0,2)$ and has a similar classification accuracy as $\mathcal{U}(0.25,1.75)$).
     In Figure~\ref{fig:privacy_uniform_everywhere} we see that token-consistent noise is more private compared to the non-token-consistent version.}}
    \label{fig:collaborative_inference}
\end{figure}

%\caption{{\textbf{Non-homeomorphic mappings}. Multiplying every element of the feature map $\mathcal{\Tilde{Z}}^{FC_1}_{i}$ with independent noise (e.g. Dropout~\cite{srivastava2014dropout}) is a non-homeomorphic operation, which does not preserve the privacy as well as homeomorphic stochastic layers. Dropout in~\ref{fig:privacy_dropout} matches the variance of $\mathcal{U}(0,2)$ and has a similar classification accuracy as stochastic layers with $\mathcal{U}(0.25,1.75)$. In~\ref{fig:privacy_uniform_everywhere} we compare our stochastic layers to applying independent noise everywhere.}}

%% file: figures/paper/privacy_split_learning.tex
\begin{figure}[t!]
    \centering
    \captionsetup{font=small}
    \begin{subfigure}[b]{0.23\columnwidth}
        \centering
        \includegraphics[width=\columnwidth]{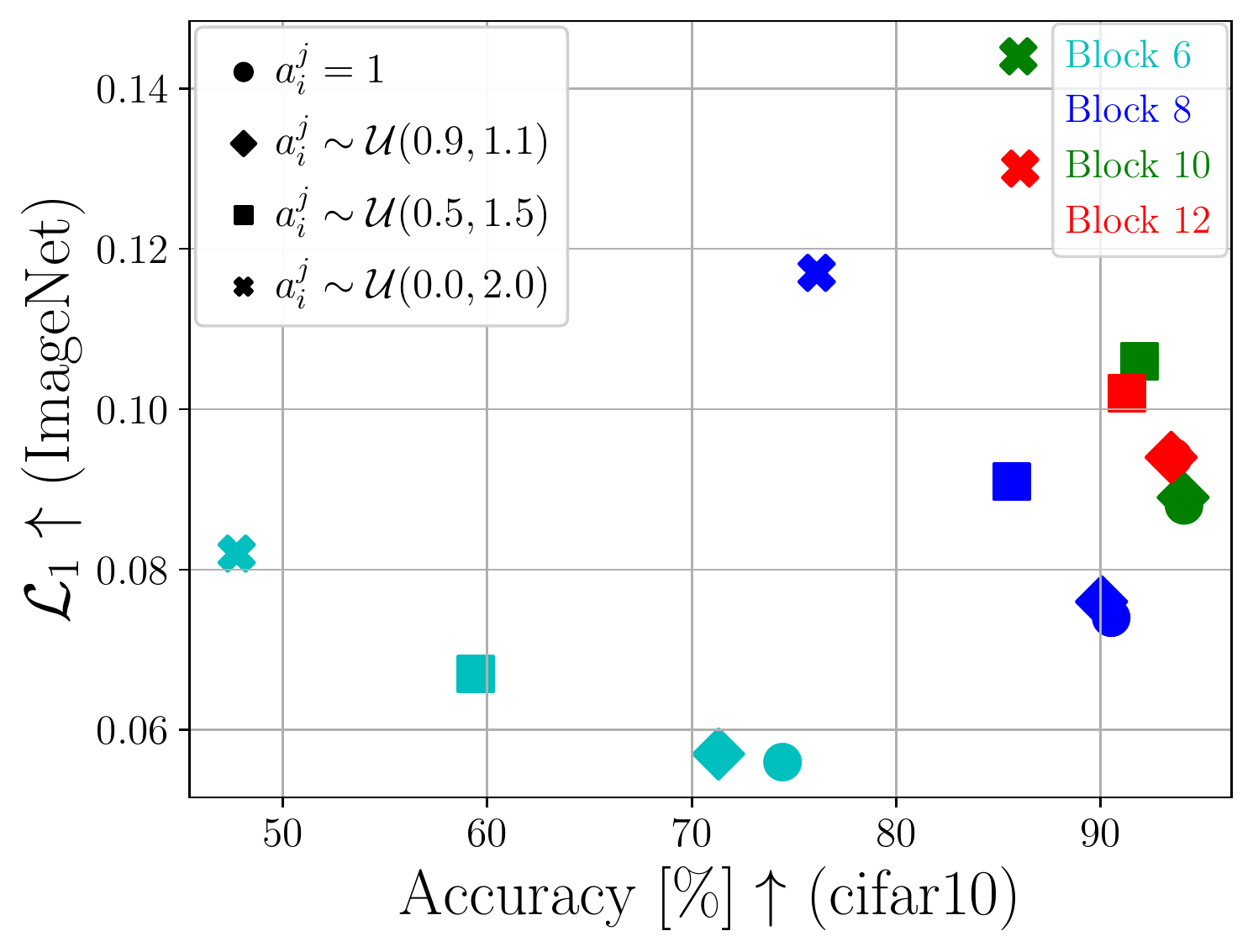}
        \caption{}
    \end{subfigure}
    \hfill    
    \begin{subfigure}[b]{0.23\columnwidth}
        \centering
        \includegraphics[width=\columnwidth]{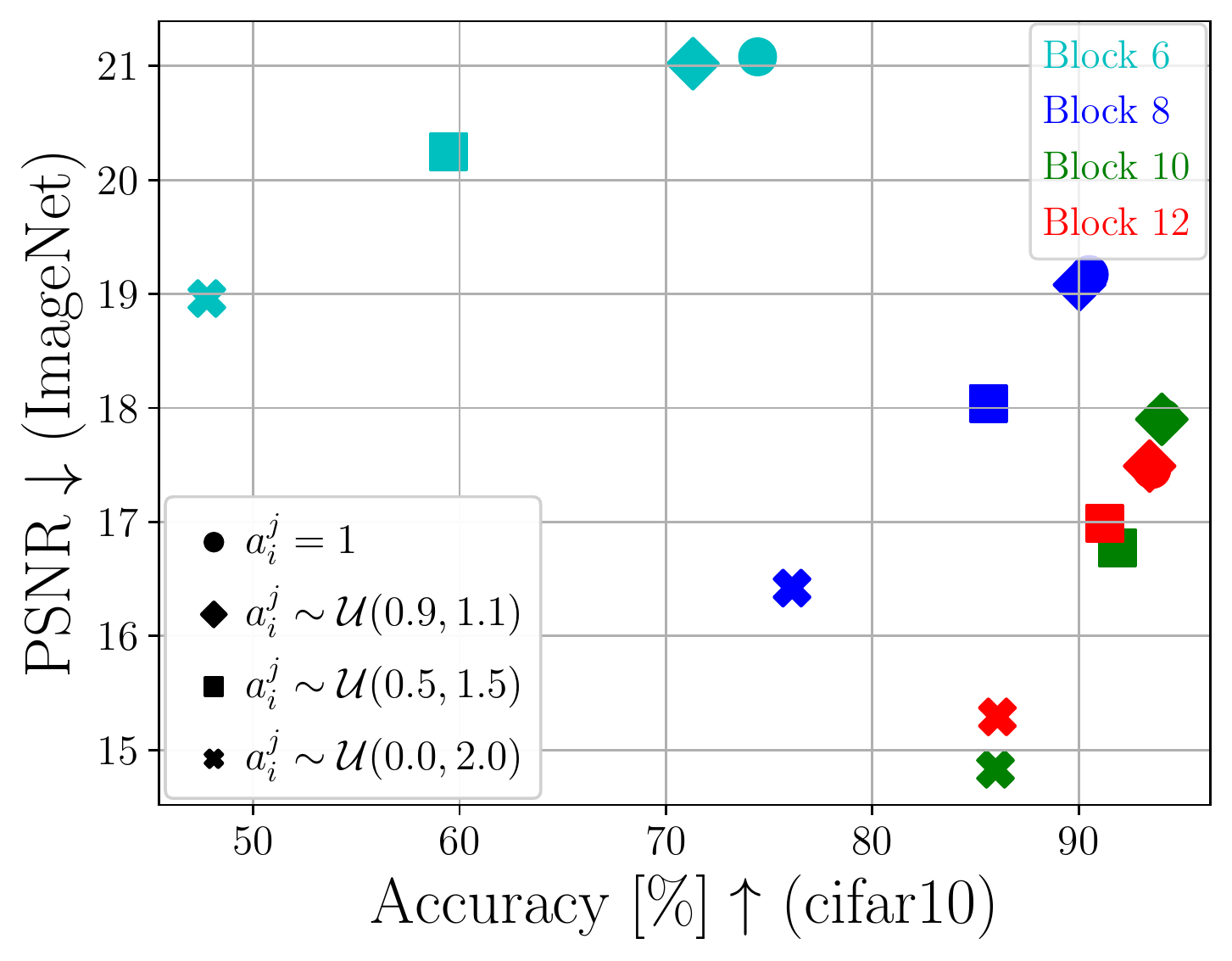}
        \caption{}
    \end{subfigure}
    %\vskip\baselineskip
    \begin{subfigure}[b]{0.23\columnwidth}
        \centering
        \includegraphics[width=\columnwidth]{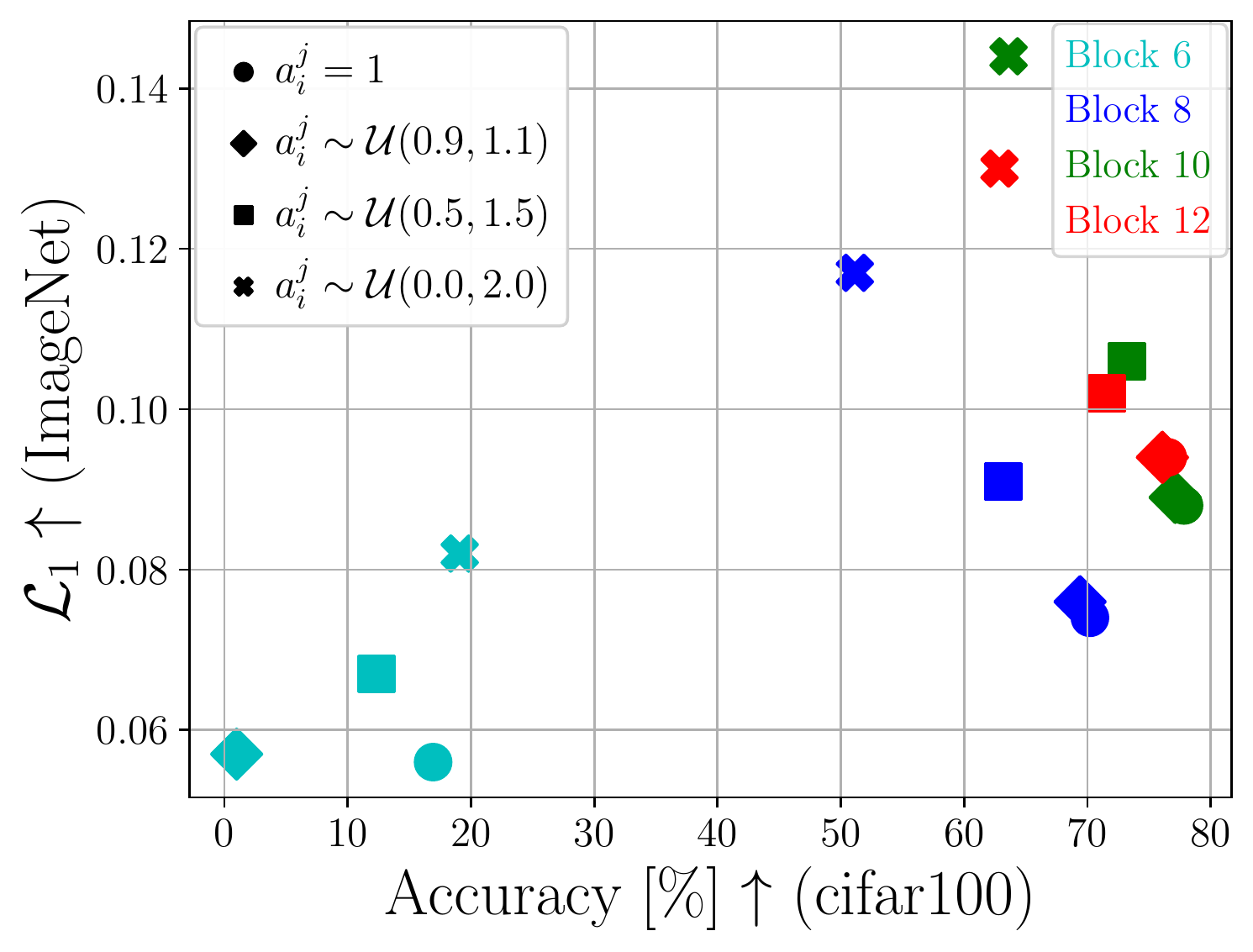}
        \caption{}
    \end{subfigure}
    \hfill    
    \begin{subfigure}[b]{0.23\columnwidth}
        \centering
        \includegraphics[width=\columnwidth]{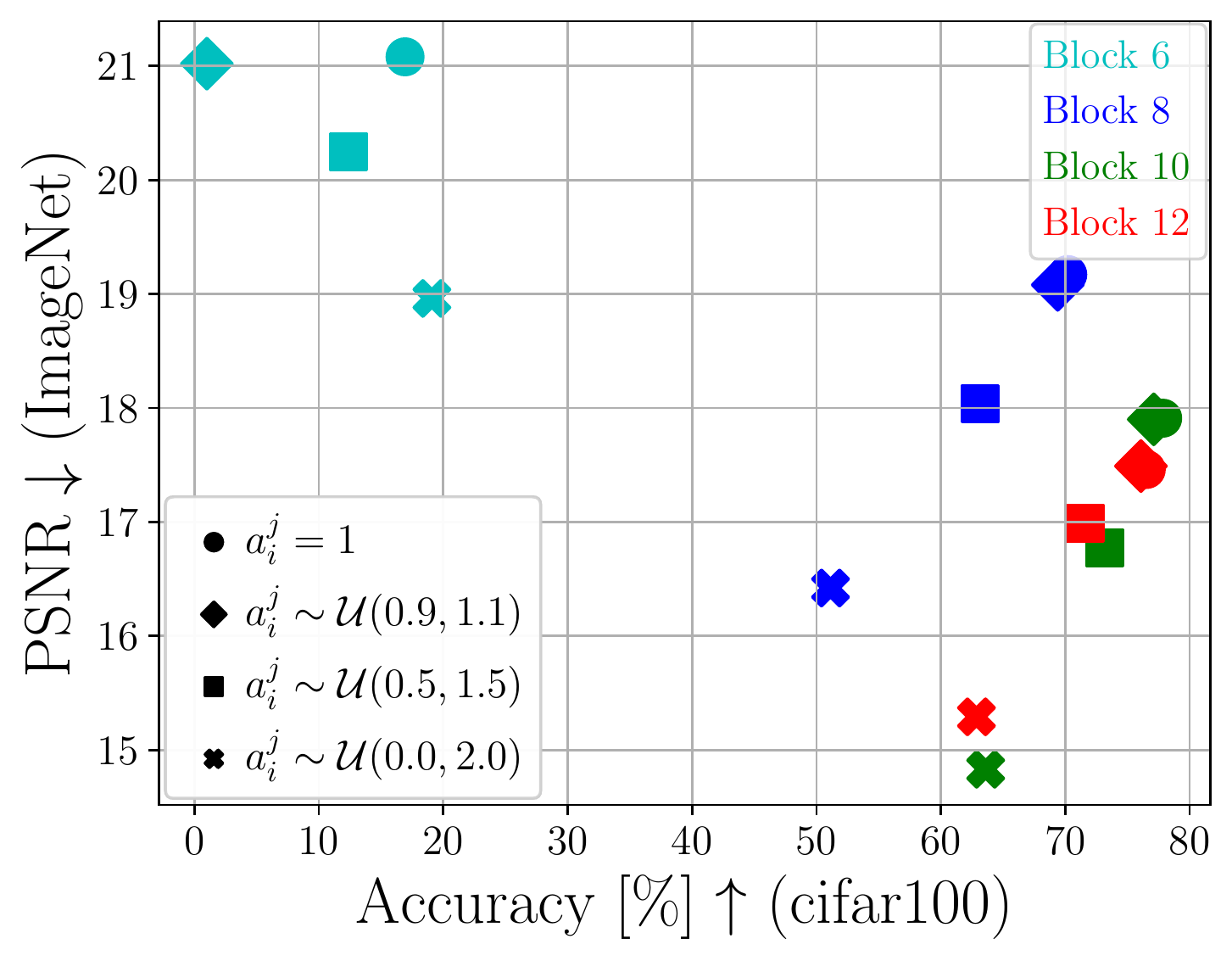}
        \caption{}
    \end{subfigure}
    \vspace{-5pt}
    \caption{{\textbf{Privacy in collaborative learning}. Stochastic layers offer privacy and performance trade-offs whose operating point can be chosen in accordance to the use-case.}}
    \label{fig:privacy_split_learning}
\end{figure}

%% file: tables/paper/transfer_learning.tex
\begin{table}[t!]
\centering
\tableFont
\captionsetup{font=small}
\caption{{\textbf{Transfer learning on 12 different tasks}. 
%The transfer learning capabilities are not negatively impacted by our stochastic layers. We even observe slight improvements for some datasets.
The token-consistent stochastic layers retain transferability across different noise levels. }}
%\begin{adjustbox}{\columnwidth}
%\renewcommand\arraystretch{1}
\resizebox{\columnwidth}{!}{
\begin{tabular}{|l|l|c|c|c|c|c|c|c|c|c|c|c|c|}
\hline

\multicolumn{2}{|c|}{} &
Aircraft & 
Birdsnap & 
\makecell{CIFAR\\10} & 
\makecell{CIFAR\\100} & 
\makecell{Caltech\\101} & 
\makecell{Caltech\\256} & 
Cars & 
DTD & 
Flowers & 
Food & 
Pets & 
\makecell{SUN\\397} \\ %\hline

\multicolumn{2}{|c|}{} &
\cite{maji13fine-grained} & 
\cite{Berg2014BirdsnapLF} & 
\cite{Krizhevsky2009LearningML} & 
\cite{Krizhevsky2009LearningML} & 
\cite{FeiFei2004LearningGV} & 
\cite{Griffin2007Caltech256OC} & 
\cite{Krause2013CollectingAL} & 
\cite{Cimpoi2014DescribingTI} & 
\cite{Nilsback2008AutomatedFC} & 
\cite{Bossard2014Food101M} & 
\cite{Parkhi2012CatsAD} & 
\cite{Xiao2010SUNDL} \\ \hline
 
\multirow{4}{*}{\makecell{Pre-\\training}} &
no noise &
$72.34\%$ &
$70.41\%$ &
$98.11\%$ &
$87.41\%$ &
$91.89\%$ &
$83.85\%$ &
$80.76\%$ &
$73.46\%$ &
$92.39\%$ &
$87.06\%$ &
$92.83\%$ &
$62.34\%$ \\ \cline{2-14}

 &
$\Delta=0.1$ &
$72.52\%$ &
$69.97\%$ &
$98.36\%$ &
$87.93\%$ &
$91.55\%$ &
$83.87\%$ &
$82.13\%$ &
$73.72\%$ &
$94.02\%$ &
$87.45\%$ &
$93.10\%$ &
$62.35\%$ \\ \cline{2-14}

 &
$\Delta=0.5$ &
$71.50\%$ &
$69.58\%$ &
$98.13\%$ &
$87.65\%$ &
$91.89\%$ &
$84.07\%$ &
$81.52\%$ &
$73.94\%$ &
$93.69\%$ &
$87.10\%$ &
$93.16\%$ &
$62.79\%$ \\ \cline{2-14}

 &
$\Delta=1.0$ &
$72.40\%$ &
$70.00\%$ &
$98.29\%$ &
$88.04\%$ &
$91.85\%$ &
$83.99\%$ &
$81.99\%$ &
$73.94\%$ &
$93.61\%$ &
$87.29\%$ &
$93.08\%$ &
$62.36\%$ \\ \hline

% Wins          & --  & $10/12$ & $10/12$  & $10/12$ \\ \hline 
\end{tabular}
}
\label{table:transfer_learning}
\end{table}

%% file: latex/paper/06_conclusion.tex
\section{Conclusion}
\label{sec:conclusion}
In this work, we investigate the role of stochasticity in vision transformers. By interjecting token-consistent stochastic layers during both training and inference, we transform the feature activations of each multilayer perceptron, while preserving the topological structure of the activations. This transformation offers the robustness and privacy of visual features, while retaining the original predictive performance. 
We demonstrated the utility of our features for the applications of adversarial robustness, network calibration, feature privacy and transfer learning.
From our experiments we conclude that our token-consistent stochastic layers are well behaved and our visual features offer exciting results on those tasks.

% %\vspace{1em}
% \noindent\textbf{Ethical and Societal Impact.}
% This work is concerned with the learning of improved visual features by noise injection. Even though these features offer better resilience on privacy and adversarial attacks, they do not meet a certifiable standard as of now. Therefore they should not be used in a production system without additional layers of safety measures, such as encryption or humans-in-the-loop.
% Although our method allows for an improved estimation of prediction confidence overall, we did not investigate the behaviour with regards to specific subgroups of classes or people. This needs to be carefully analysed with appropriate datasets to mitigate discriminatory actions by a decision-making system based on our algorithm.

%% file: latex/supplementary/0_supplementary_structure.tex
%\section{Supplementary Structure}

In this supplementary document, we first give additional details on our implementation and experimental setup, which are contained in Section~\ref{sec:supp_implementation_details}. 
In Section~\ref{sec:supp_further_analysis}, we complement the results from the main paper by showing additional comparisons and studies regarding the training process, privacy preservation, adversarial robustness, confidence calibration, and transfer learning. 
We round off the document by a discussion of the effect and rationale of our design choices, as well as several aspects of our experimental findings, which can be found in Section~\ref{sec:supp_discussion}.

%% file: latex/supplementary/1_implementation_details.tex
\section{Additional implementation details}
\label{sec:supp_implementation_details}

\subsection{Monte Carlo Inference}
%\label{ssec:monte_carlo_inference}
It has been demonstrated in the literature that the posterior distribution on the network weights $w$ can be estimated using monte-carlo sampling in the context of dropout~\cite{Gal2016DropoutAA,Kendall2017WhatUD,Gal2015BayesianCN}. This way, predictions are integrated implicitly over the posterior distribution of the network weights $q(w)$ as,
\begin{equation}
\label{eq:mc-inference}
    p(y|x) \approx \int p(y | x, w) q(w) dw \approx \frac{1}{T} \sum_{t=1}^{T} p(y| x, \hat{w}_t),
\end{equation}
approximated by T samples $\hat{w}_t$~\cite{Gal2015BayesianCN}, generated by applying dropout during inference.
The posterior distribution $p(y|x)$ also contains information about model uncertainty.
Since our stochastic layers naturally offer such sampling for vision transformers, our goal is to investigate their effect on network calibration, as well as other applications, when performing inference with~\eqref{eq:mc-inference}.

\subsection{Expectation over Transformation (EoT) Adversarial Attack}
When dealing with stochasticity inside the network, Athalye et al.~\cite{Athalye2018ObfuscatedGG} demonstrated that for the class of randomness-based defenses, an expectation-over-transformation (EOT) attack is more effective, because of common gradient obfuscation. This can be viewed as using PGD~\cite{Madry2018TowardsDL}
% \ifx\FileIsMerged\undefined
% (Equation (16) from the main paper) 
% \else
% (Equation~\eqref{eq:pgd} from the main paper) 
% \fi
with the proxy gradient,
\begin{equation}
\label{eq:pgd_expectation}
     \mathbb{E}_{q(w)} \left[\nabla_x \mathcal{L}(f(\hat{x}^{k-1}, w), l)\right] \approx \frac{1}{T} \sum_{t=1}^{T} \nabla_x \mathcal{L}(f(\hat{x}^{k-1}, w_t), l) ,
\end{equation}
where $q(w)$ represents the distribution of the noise ${w\sim q(w)}$ injected into the randomized classifier $f(x,w)$.

\subsection{Training our stochastic layers}
When fine-tuning our network to adapt to the stochastic layers with $a^{j} \sim \mathcal{U}(1-\Delta,1+\Delta)$ 
\ifx\FileIsMerged\undefined
(from Equation (8) of the main paper)
\else
(from Equation~\eqref{eq:noise_injection} of the main paper)
\fi 
, we gradually adjust $\Delta$ from $0$ to its final value during the first third of the training epochs. When we use Dropout as a baseline, while matching the variance of $a^{j} \sim \mathcal{U}(1-\Delta,1+\Delta)$, we also increase the drop probability from $0$ to its final value in this manner. The same goes for the baseline which independently draws $a^{j} \sim \mathcal{U}(1-\Delta,1+\Delta)$ for each token -- non-token-consistent.
Also, in the experiment with adversarial training 
\ifx\FileIsMerged\undefined
(see Section 5.3 of the main paper)
\else
(see Section~\ref{sec:adversarial_robustness} of the main paper)
\fi 
and in experiments when we train from scratch 
\ifx\FileIsMerged\undefined
(see Table 1c of the main pape),
\else
(see Table~\ref{table:imnet_100_scratch_accuracy} of the main paper),
\fi 
we use the same $\Delta$ scheduling rule.

\subsection{Collaborative inference}
When performing the experiments mentioned in 
\ifx\FileIsMerged\undefined
Figure~4 of the main paper,
\else
Figure~\ref{fig:collaborative_inference} of the main paper,
\fi 
 we train the reconstruction decoder for $100$ epochs with a batch size of $32$. The weights of the client network $\Phi_1(x)$, which produces the feature maps, are frozen. The stochasticity remains turned on during both training and testing.

\subsection{Collaborative learning}
When performing the experiments mentioned in  
\ifx\FileIsMerged\undefined
Figure~5 of the main paper,
\else
Figure~\ref{fig:privacy_split_learning} of the main paper,
\fi 
 the client network $\Phi_1(x)$ is the part of the transformer which produces the feature map and its weights are frozen. The server network $\Phi_2(x)$ is a small classifier with $2$ fully connected layers, followed by a softmax. $\Phi_2(x)$ is trained for a $100$ epochs, with the learning rate of $0.0001$ and the AdamW optimizer. The learning rate is divided by $10$ after $50$ epochs. The stochasticity remains turned on during both training and inference.

%% file: latex/supplementary/2_further_analysis.tex
\section{Further analysis}
\label{sec:supp_further_analysis}

In this section, we perform a further analysis of the experiments presented in 
\ifx\FileIsMerged\undefined
Section 5 of the main paper.
\else
Section~\ref{sec:experiments} of the main paper.
\fi

\input{tables/supplementary/classification_image_net_1K}
Table~\ref{table:supp_imnet_1K_accuracy} is an extended version of
\ifx\FileIsMerged\undefined
Table~1a of the main paper.
\else
Table~\ref{table:imnet_1k_accuracy_thorough} of the main paper.
\fi
We see that there is no significant drop in accuracy when using our stochastic layers. To a brief discussion on the slightly better performance of regular dropout and non-token-consistent stochastic layers, please look at Section~\ref{sec:supp_dropout_diff} of this supplementary material and 
\ifx\FileIsMerged\undefined
Section~5.1 of the main paper.
\else
Section~\ref{sec:experiments_stochastic_layers} of the main paper.
\fi

\input{tables/supplementary/adv_robustness_raw}

Next, we further analyse the problem discussed in 
\ifx\FileIsMerged\undefined
Section 5.3 of the main paper.
\else
Section~\ref{sec:adversarial_robustness} of the main paper.
\fi
In Table~\ref{tab:supp_adv_robustness_raw} we see that our token-consistent stochastic layers achieve some level of adversarial robustness without using adversarial training. This is not the case for the regular transformer.
% Furthermore, Table~\ref{table:supp_adv_robustness} complements 
% \ifx\FileIsMerged\undefined
% Table 2b of the main paper.
% \else
% Table~\ref{tab:supp_adv_robustness_raw} of the main paper.
% \fi
%\input{tables/supplementary/adv_robustness}

\input{figures/supplementary/reconstruction_visualization}

In the following, we further analyse the problem of collaborative inference discussed in 
\ifx\FileIsMerged\undefined
Section 5.4 of the main paper.
\else
Section~\ref{ssec:private_features} of the main paper.
\fi
In Figure~\ref{fig:supp_reconstruction} we see examples of image reconstruction from the feature map of block $7$ of the transformer. 
The first row depicts the original image. The second row depicts images reconstructed from the feature map of a deterministic network. Finally, the third row depicts reconstructions from our stochastic network. In particular the facial details are more degraded by our stochastic layers.
% For a brief discussion of these results, please look at Section~\ref{sec:supp_discuss_privacy}.

\input{tables/supplementary/collaborative_learning_before_after_act}

Now, we further analyse the problem of collaborative learning discussed in 
\ifx\FileIsMerged\undefined
Section 5.4 of the main paper.
\else
Section~\ref{ssec:private_features} of the main paper.
\fi
In Table~\ref{table:supp_collaborative_learning_before_after_act} we observe the results of learning a classifier on CIFAR-10 with features before and after the activation function. We observe that using features before the activation function to learn a classifier achieves better performance than when using features after the activation. This justifies the choice of using the features before the ReLU activation, in collaborative learning experiments.

\input{tables/supplementary/transfer_learning}

Finally, we further analyse the problem of transfer learning from 
\ifx\FileIsMerged\undefined
Section 5.5. of the main paper.
\else
Section~\ref{sec:transfer_learning} of the main paper.
\fi
In Table~\ref{table:supp_transfer_learning} we observe that our token-consistent stochastic layers are able to retain their transfer learning performance and sometimes even exceed the deterministic baseline.

\input{figures/supplementary/topology_1}
\input{figures/supplementary/topology_3}

In Figures~\ref{fig:supp_topology1_bottleneck0},~\ref{fig:supp_topology1_bottleneck1}~and~\ref{fig:supp_topology1_bottleneck2} we see histograms of barcode distances~\cite{guss2018characterizing,chazal2021introduction} of feature maps before and after the stochasticity is injected. The feature maps are obtained from the last block of the transformer.  Histograms are constructed from $1000$ randomly chosen input images from the validation set. The stochasticity level is $\Delta=0.5$. In Figures~\ref{fig:supp_topology1_dropout},~\ref{fig:supp_topology1_univorm_everywhere}~and~\ref{fig:supp_topology1_ours} we can see the barcode persistence diagrams for a randomly chosen input image for dropout, non-token-consistent and token-consistent stochastic layers, respectively.  
%We conclude that our stochastic layers preserve the topological structure of token features like it was mentioned in Section 3 of the main paper.
Based on small distances of our method before and after injecting stochasticity, we conclude that the topological structure of token features is preserved.
For a brief discussion on this experiment, please see Section~\ref{sec:supp_discuss_topology}.

In Figure~\ref{fig:supp_topology_3} we see histograms of average barcode distances between $4$ samples for our stochastic network and the deterministic baseline. Stochasticity with the level of $\Delta=0.5$ is turned on for both the stochastic network and the determenistic baseline as well. We take the feature map from the last block of the transformer, right before the stochasticity is injected, and sample $4$ inference forward passes. Histograms are constructed from $1000$ randomly chosen input images from the validation set. 
%We observe that feture maps of our network have a more consistent topology compared to the determenistic baseline, when stochasticity is injected.
We observe that when a stochasticity is injected, feature maps of our stochastic network have a more consistent topology (lower barcode distances) compared to the deterministic baseline. For a brief discussion of this experiment, please see Section~\ref{sec:supp_discuss_topology}.

%% file: tables/supplementary/classification_image_net_1K.tex
\begin{table}[t!]
\centering
\tableFont
\captionsetup{font=small}
\caption{{\textbf{ImageNet-1k classification accuracy}. There is no significant drop in accuracy when using our token-consistent stochastic layers.}}
%\begin{adjustbox}{\columnwidth}
%\renewcommand\arraystretch{1}
%\resizebox{\columnwidth}{!}{
\begin{tabular}{|l|l|c|c|c|}
\hline
\multicolumn{2}{|c|}{} & \multicolumn{3}{c|}{Accuracy$\uparrow$}  \\ \cline{3-5}
\multicolumn{2}{|c|}{} & No noise & $N=1$ & $N=50$         \\ \hline \hline

\multicolumn{2}{|l|}{DeiT-S}  & $79.61\%$ & $79.61\%$ & $79.61\%$  \\ \hline \hline
\multicolumn{2}{|l|}{DeiT-S + token-consistent  ($\Delta=0.05$)}                       & $79.79\%$ & $79.71\%$ & $79.81\%$  \\ \hline \hline

\multicolumn{2}{|l|}{DeiT-S + token-consistent  ($\Delta=0.1$)}                       & $79.83\%$ & $79.58\%$ & $79.86\%$  \\ \hline \hline

\multicolumn{2}{|l|}{DeiT-S + token-consistent ($\Delta=0.25$)}                       & $79.87\%$ & $79.28\%$ & $79.88\%$  \\ \hline \hline

\multirow{3}{*}{\makecell{Training \\ $\Delta=0.5$}}  
 & DeiT-S + token-consistent & $79.60\%$ & $77.43\%$ & $79.72\%$ \\ \cline{2-5}
 & DeiT-S + non-token-consistent & $79.63\%$ & $78.59\%$ & $79.81\%$ \\ \cline{2-5}
 & DeiT-S + dropout& $79.60\%$ & $78.54\%$ & $79.79\%$ \\ \hline \hline
 
\multicolumn{2}{|l|}{DeiT-S + token-consistent  ($\Delta=0.75$)}                       & $78.93\%$ & $75.46\%$ & $78.83\%$  \\ \hline \hline
 
\multirow{3}{*}{\makecell{Training \\ ($\Delta=1.0$)}} 
 & DeiT-S + token-consistent & $77.96\%$ & $72.93\%$ & $77.75\%$ \\ \cline{2-5} 
 & DeiT-S + non-token-consistent& $77.99\%$ & $75.85\%$ & $78.63\%$ \\ \cline{2-5} 
 & DeiT-S + dropout & $77.93\%$ & $76.18\%$ & $78.74\%$ \\ \hline

\end{tabular}
%}
\label{table:supp_imnet_1K_accuracy}
\end{table}

%% file: tables/supplementary/adv_robustness_raw.tex
\begin{table}[t!]
\centering
\tableFont
\captionsetup{font=small}
\caption{{\textbf{Adversarial robustness, without adversarial training.}
We see that using token-consistent stochastic layers achieves some level of robustness without adversarial training. This is not the case for the regular transformer. We also observe that the EoT attack does not fully penetrate the stochastic defense in this case.
}}
%\resizebox{\columnwidth}{!}{
%\renewcommand\arraystretch{1}
\centering
%\resizebox{0.8\columnwidth}{!}{
\begin{tabular}{|c|l||c|c|c|c|c|}
\hline 
\multicolumn{2}{|c||}{} & \multicolumn{5}{c|}{Accuracy $\uparrow$ (Without adversarial training)} \\ \cline{3-7}

\multicolumn{2}{|c||}{} 
 & \multirow{3}{*}{\makecell{Clean \\ samples}}
 & \multicolumn{4}{c|}{\makecell{PDG$-10$ attack }} 
 \\ \cline{4-7}  
\multicolumn{2}{|c||}{} 
 & 
 & \multicolumn{2}{c|}{\makecell{$\epsilon=\frac{2}{255}$}}
 & \multicolumn{2}{c|}{\makecell{$\epsilon=\frac{4}{255}$}}
 \\ \cline{4-7} 
\multicolumn{2}{|c||}{} 
 & 
 & \makecell{}
 & \makecell{EoT}
 & \makecell{}
 & \makecell{EoT}
 \\ \hline   
 
\multicolumn{2}{|c||}{DeiT-S} & $\textbf{79.61\%}$ & $0.43\%$ & $0.43\%$ & $0.01\%$ & $0.01\%$ \\ \hline 

%\multirow{4}{*}{\makecell{Training \\ $\Delta$}}

%  %& \multirow{2}{*}{$0.1$}
%  %& $N=1$ & $\%$ & $\%$ & \\ \cline{3-8} 
%  %& & $N=50$ & $\%$ & $\%$ \\ \cline{2-8}

 \multirow{2}{*}{\makecell{ DeiT-S + ours \\ ($\Delta=0.5$)}}
 & $N=1$ & $77.43\%$ & $5.25\%$ & $1.97\%$ & $1.41\%$ & $0.23\%$ \\ \cline{2-7} 
 & $N=50$ & $79.72\%$ & $5.25\%$ & $1.85\%$ & $1.37\%$ & $0.23\%$ \\ \hline
 
 \multirow{2}{*}{\makecell{ DeiT-S + ours\\ ($\Delta=1.0$)}}
 & $N=1$ & $72.93\%$ & $12.7\%$ & $\textbf{4.80\%}$  & $4.77\%$ & $1.06\%$ \\ \cline{2-7}  
 & $N=50$ & $77.75\%$ & $\textbf{12.36\%}$ & $4.77\%$ & $\textbf{4.76\%}$ & $\textbf{1.07\%}$  \\ \hline
\end{tabular}
%}
\label{tab:supp_adv_robustness_raw}
%}
\end{table}

%% file: figures/supplementary/reconstruction_visualization.tex
\begin{figure*}[t]
  \centering
  \captionsetup{font=small}
  \tiny
  \setlength{\tabcolsep}{1pt}
  \vspace*{2pt}
  \newcommand{\sz}{0.11}           % figure size
  \newcommand{\cgs}{\hspace{3pt}}   % define spacing between column groups
  \begin{tabular}{cccccccc}
    \multirow{1}{*}[27pt]{\rotatebox{90}{Original}}
    \includegraphics[width=\sz\columnwidth]{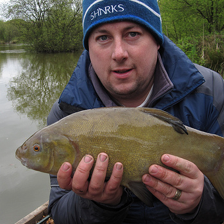} & 
    \includegraphics[width=\sz\columnwidth]{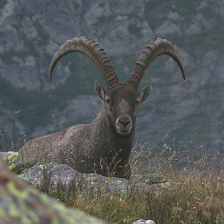} & 
    \includegraphics[width=\sz\columnwidth]{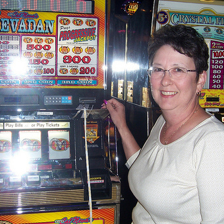} &
    \includegraphics[width=\sz\columnwidth]{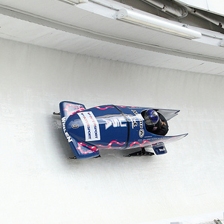} & 
    \includegraphics[width=\sz\columnwidth]{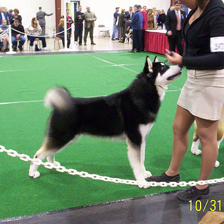} &
    \includegraphics[width=\sz\columnwidth]{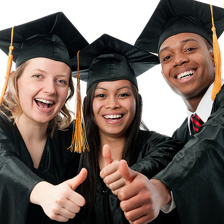} &
    \includegraphics[width=\sz\columnwidth]{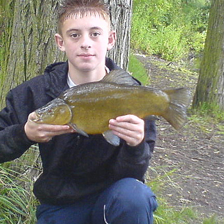} & 
    \includegraphics[width=\sz\columnwidth]{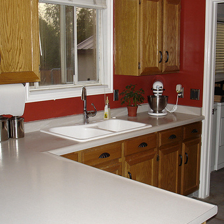} \\
    
    \multirow{1}{*}[36pt]{\rotatebox{90}{Deterministic}}
    \includegraphics[width=\sz\columnwidth]{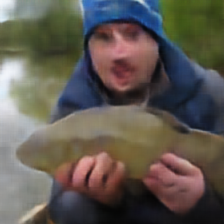} & 
    \includegraphics[width=\sz\columnwidth]{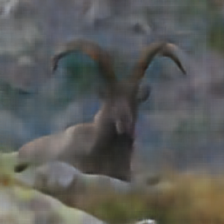} & 
    \includegraphics[width=\sz\columnwidth]{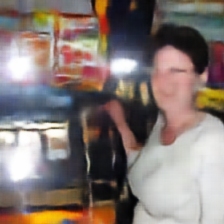} &  
    \includegraphics[width=\sz\columnwidth]{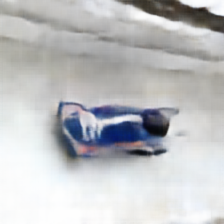} & 
    \includegraphics[width=\sz\columnwidth]{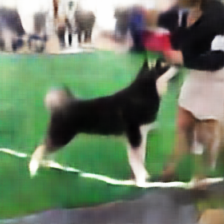} &
    \includegraphics[width=\sz\columnwidth]{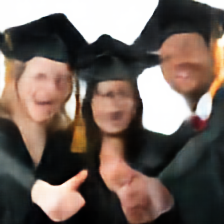} &
    \includegraphics[width=\sz\columnwidth]{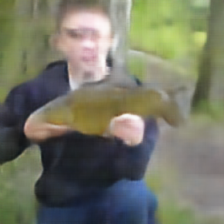} & 
    \includegraphics[width=\sz\columnwidth]{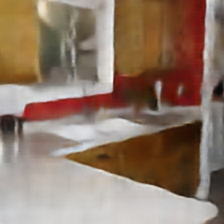}  \\  
    
    \multirow{1}{*}[32pt]{\rotatebox{90}{$a^{j} \sim \mathcal{U}(0, 4)$}}
    \includegraphics[width=\sz\columnwidth]{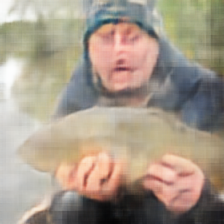} & 
    \includegraphics[width=\sz\columnwidth]{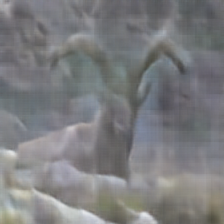} & 
    \includegraphics[width=\sz\columnwidth]{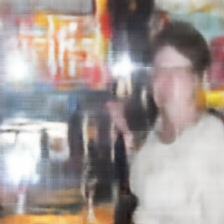} &
    \includegraphics[width=\sz\columnwidth]{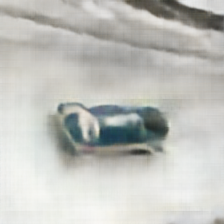} &  
    \includegraphics[width=\sz\columnwidth]{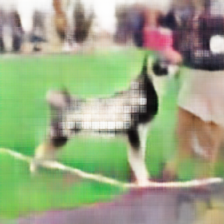} &
    \includegraphics[width=\sz\columnwidth]{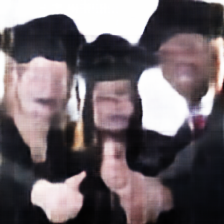} &
    \includegraphics[width=\sz\columnwidth]{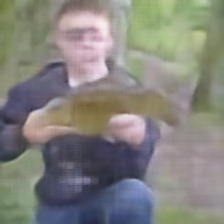} & 
    \includegraphics[width=\sz\columnwidth]{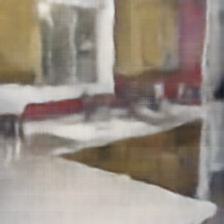}  \\

  \end{tabular}
 \caption{\textbf{Privacy preservation in collaborative inference.} Reconstructed images from block $7$ of the transformer with and without our token-consistent stochastic layers. First row depicts the original image. The second row depicts images reconstructed from the feature map of a deterministic transformer. Finally, the third row depicts reconstructions from our stochastic network. Our token-consistent stochastic layers preserve more privacy, in particular in the facial details.}
    \label{fig:supp_reconstruction}
\end{figure*}

%% file: tables/supplementary/collaborative_learning_before_after_act.tex
\begin{table}[t!]
\centering
\tableFont
\captionsetup{font=small}
\caption{{\textbf{Learning a classifier on CIFAR-10 with features before and after the activation function}. Using features before the activation function to learn a classifier achieves better performance than when using features after the activation.}}
%\begin{adjustbox}{\columnwidth}
%\renewcommand\arraystretch{1}
%\resizebox{\columnwidth}{!}{
\begin{tabular}{|l|c|c|c|c|c|c|c|c|}
\hline
 & \multicolumn{8}{c|}{Accuracy$\uparrow$}  \\ \cline{2-9}
 & \multicolumn{2}{c|}{$\Delta=0$} 
 & \multicolumn{2}{c|}{$\Delta=0.1$} 
 & \multicolumn{2}{c|}{$\Delta=0.5$}
 & \multicolumn{2}{c|}{$\Delta=1.0$}\\ \hline
 & \makecell{Before \\ activation}
 & \makecell{After \\ activation}
 & \makecell{Before \\ activation}
 & \makecell{After \\ activation}
 & \makecell{Before \\ activation}
 & \makecell{After \\ activation}
 & \makecell{Before \\ activation}
 & \makecell{After \\ activation} \\ \hline \hline

Block $6$  & $74.44\%$ & $75.93\%$ & $71.31\%$ & $74.54\%$ & $59.45\%$ & $62.79\%$ & $47.75\%$ & $49.09\%$ \\ \hline 

Block $8$  & $90.51\%$ & $81.27\%$ & $90.04\%$ & $81.67\%$ & $85.64\%$ & $77.18\%$ & $76.11\%$ & $68.07\%$ \\ \hline 

Block $10$  & $94.07\%$ & $87.48\%$ & $94.03\%$ & $86.99\%$ & $91.9\%$ & $84.37\%$ & $85.97\%$ & $77.90\%$ \\ \hline 

Block $12$  & $93.54\%$ & $88.98\%$ & $93.44\%$ & $89.19\%$ & $91.27\%$ & $86.68\%$ & $86.06\%$ & $80.55\%$ \\ \hline 

\end{tabular}
%}
\label{table:supp_collaborative_learning_before_after_act}
\end{table}

%% file: tables/supplementary/transfer_learning.tex
\begin{table}[t!]
\centering
\tableFont
\captionsetup{font=small}
\caption{{{\textbf{Transfer learning}}. 
Transformers with our token-consistent stochastic layers are able to retain their performance and sometimes even exceed the deterministic baseline.}}
%\begin{adjustbox}{\columnwidth}
%\renewcommand\arraystretch{1}
%\resizebox{\columnwidth}{!}{
\begin{tabular}{|l|c|c|c|c|c|c|c|}
\hline
%& \multicolumn{4}{c|}{Accuracy$\uparrow$}  \\ \cline{2-5}
\multirow{3}{*}{Dataset} & \multirow{3}{*}{\makecell{Regular \\ pre-training}} & \multicolumn{6}{c|}{$\Delta$ during pre-training} \\ \cline{3-8}
 &  & \multicolumn{3}{c|}{$0.5$} & \multicolumn{3}{c|}{$1.0$} \\ \cline{3-8}
 &  & \makecell{Token-\\consistent} & \makecell{Non-\\token-\\consistent} & Dropout & \makecell{Token-\\consistent} & \makecell{Non-\\token-\\consistent} & Dropout \\ \hline \hline 

Aircraft        & $72.34\%$  & $71.50\%$ & $73.18\%$ & $72.46\%$  & $72.40\%$ & $71.71\%$  & $70.15\%$ \\ \hline 
Birdsnap        & $70.41\%$ & $69.58\%$ & $70.64\%$ & $70.33\%$  &$70.00\%$ & $70.19\%$  & $69.80\%$  \\ \hline 
CIFAR-10        & $98.11\%$ & $98.13\%$ & $98.35\%$  & $98.16\%$  & $98.29\%$ & ${98.35\%}$  & ${98.35\%}$  \\ \hline 
CIFAR-100       & $87.41\%$ & $87.65\%$ & $87.68\%$  & $87.82\%$  & ${88.04\%}$ & $87.87\%$  & $87.54\%$  \\ \hline 
Caltech-101     & $91.89\%$ & $91.89\%$ & $91.78\%$  & $91.69\%$  & $91.85\%$ & $92.21\%$  & ${92.83\%}$ \\ \hline 
Caltech-256     & $83.85\%$ & ${84.07\%}$ & $83.85\%$  & $83.78\%$  &$83.99\%$ & $83.52\%$  & $83.06\%$  \\ \hline 
Cars            & $80.76\%$ &  $81.52\%$ & $80.94\%$  & $81.54\%$  &$81.99\%$ & ${82.34\%}$  & $81.46\%$  \\ \hline  
DTD             & $73.46\%$ &  $73.94\%$ & $74.31\%$  & ${74.95\%}$  &$73.94\%$ & $72.71\%$  & $73.19\%$ \\ \hline 
Flowers         & $92.39\%$ &  ${93.69\%}$ & $93.58\%$  & $93.28\%$  & $93.61\%$ & $93.43\%$  & $92.52\%$ \\ \hline 
Food            & $87.06\%$ &  $87.10\%$ & $87.12\%$  & $87.25\%$ & ${87.29\%}$ & $86.95\%$ & $87.05\%$ \\ \hline 
Pets            & $92.83\%$ &  ${93.16\%}$ & $92.86\%$ & $92.89\%$ & $93.08\%$ & $92.89\%$ &$93.02\%$ \\ \hline
SUN397          & $62.34\%$ &  ${62.79\%}$ & $62.72\%$ & $62.29\%$ & $62.36\%$ & $62.41\%$ & $62.25\%$ \\ \hline 
%Wins          & --  & 10/12 & 10/12  & 10/12 & &\\ \hline 

\end{tabular}
%}
\label{table:supp_transfer_learning}
\end{table}

%% file: figures/supplementary/topology_1.tex
\begin{figure*}[t!]
    \centering
    \captionsetup{font=small}
    \begin{subfigure}[b]{0.32\linewidth}
        \centering
        \includegraphics[width=0.8\linewidth]{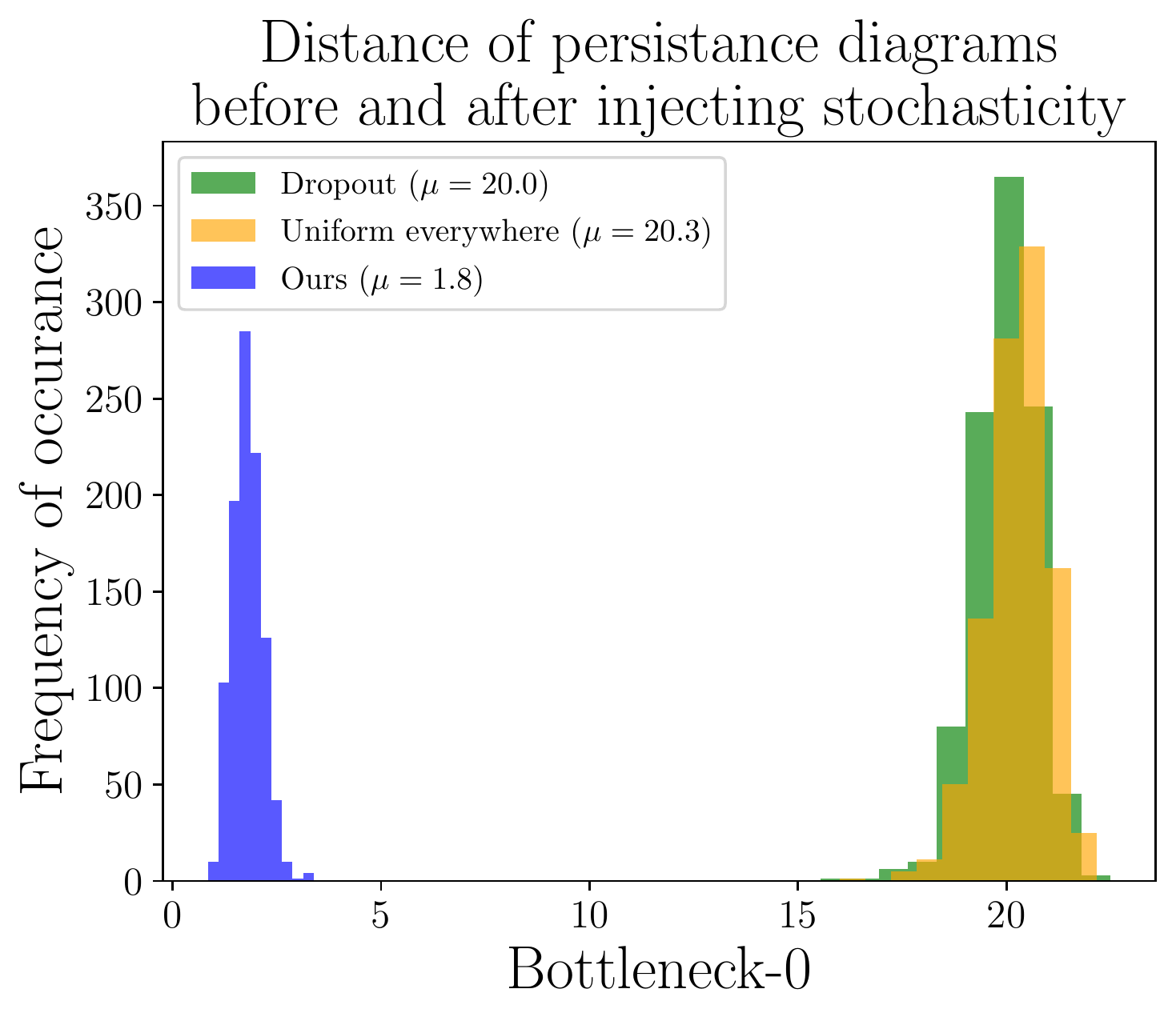}
        \caption{}
        \label{fig:supp_topology1_bottleneck0}
    \end{subfigure}
    \hfill 
    %\vskip\baselineskip
    \begin{subfigure}[b]{0.32\linewidth}
        \centering
        \includegraphics[width=0.8\linewidth]{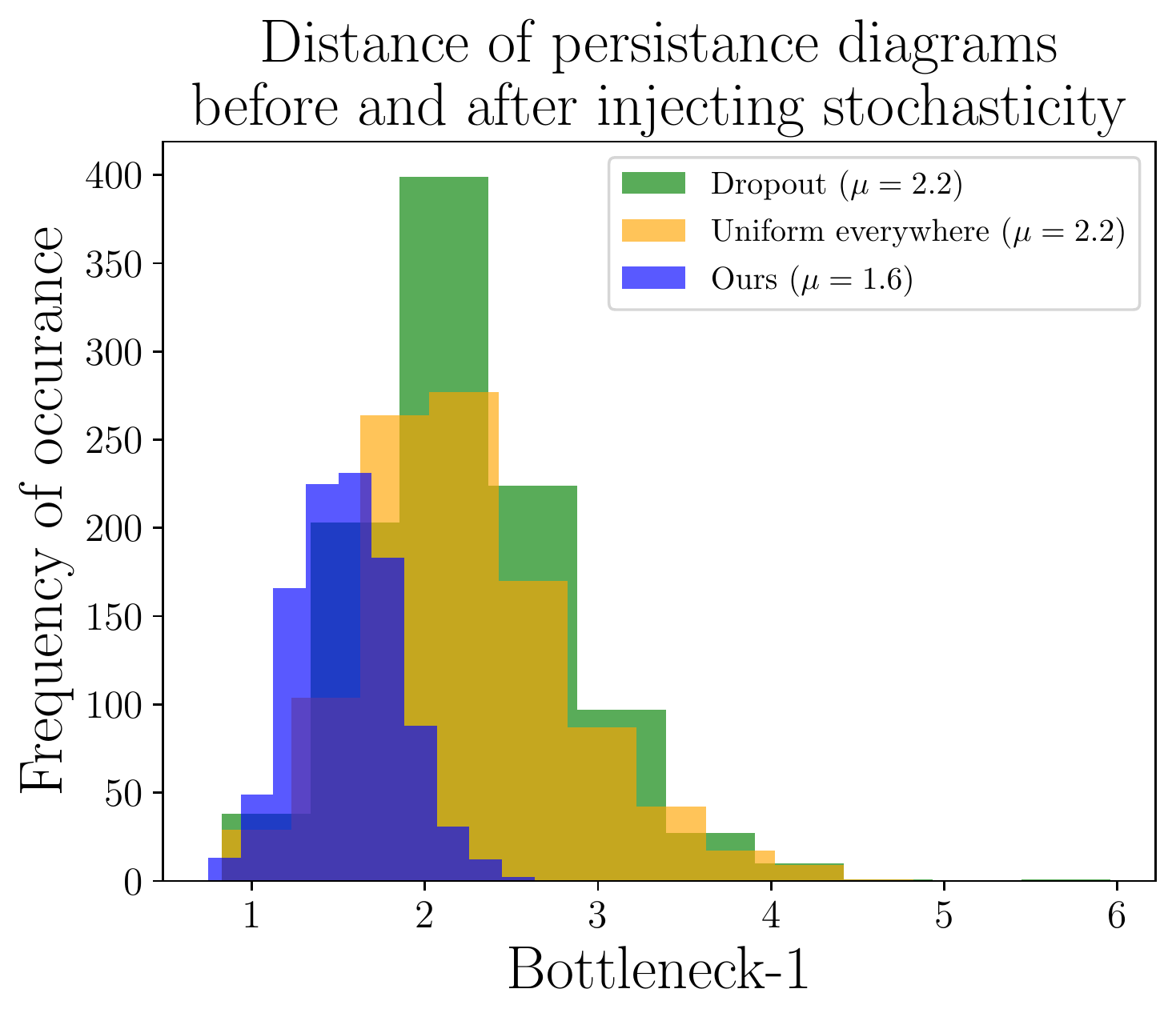}
        \caption{}
        \label{fig:supp_topology1_bottleneck1}
    \end{subfigure}
    \hfill    
    \begin{subfigure}[b]{0.32\linewidth}
        \centering
        \includegraphics[width=0.8\linewidth]{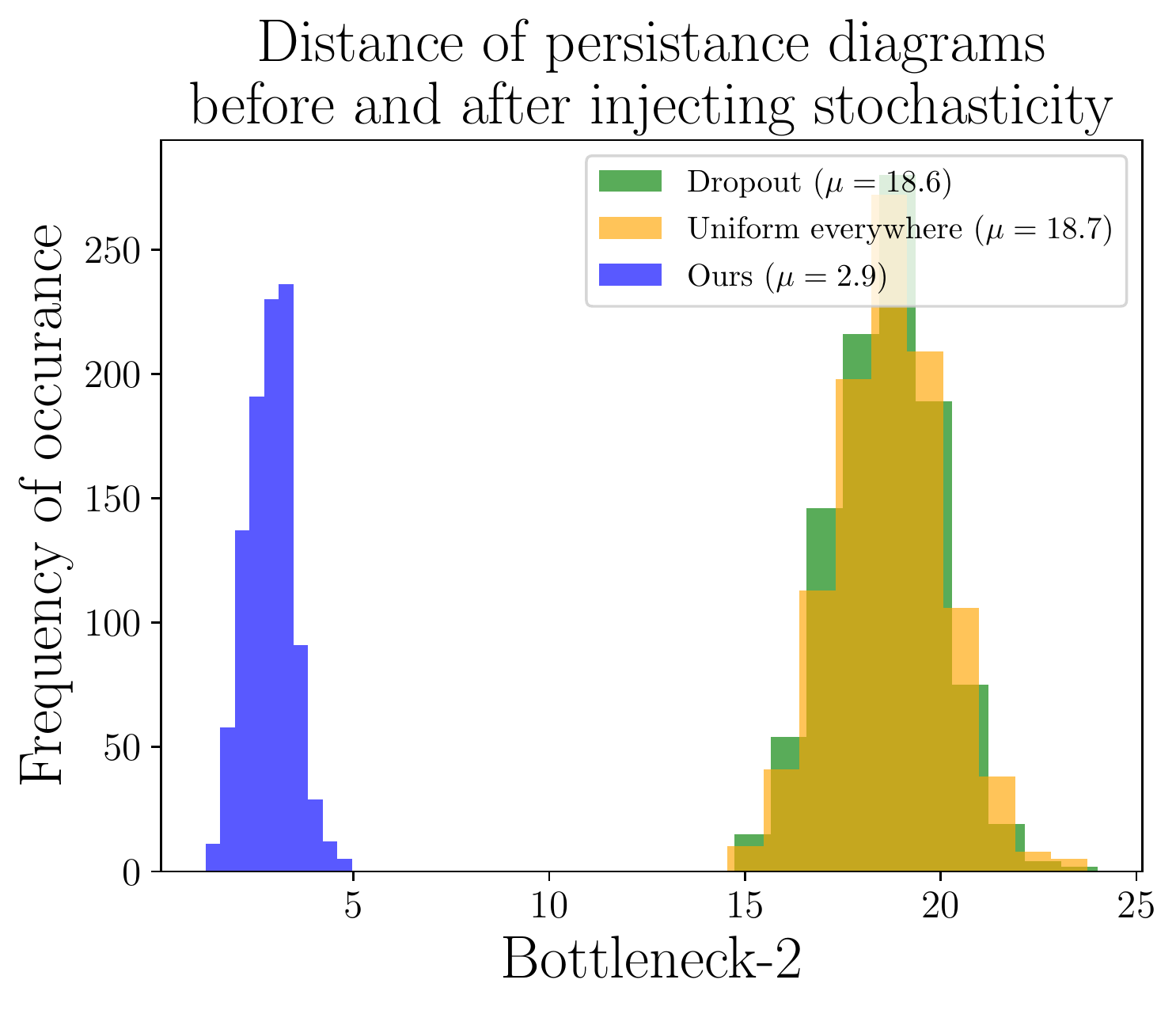}
        \caption{}
        \label{fig:supp_topology1_bottleneck2}
    \end{subfigure}
    
    \begin{subfigure}[b]{0.32\linewidth}
        \centering
        \includegraphics[width=0.8\linewidth]{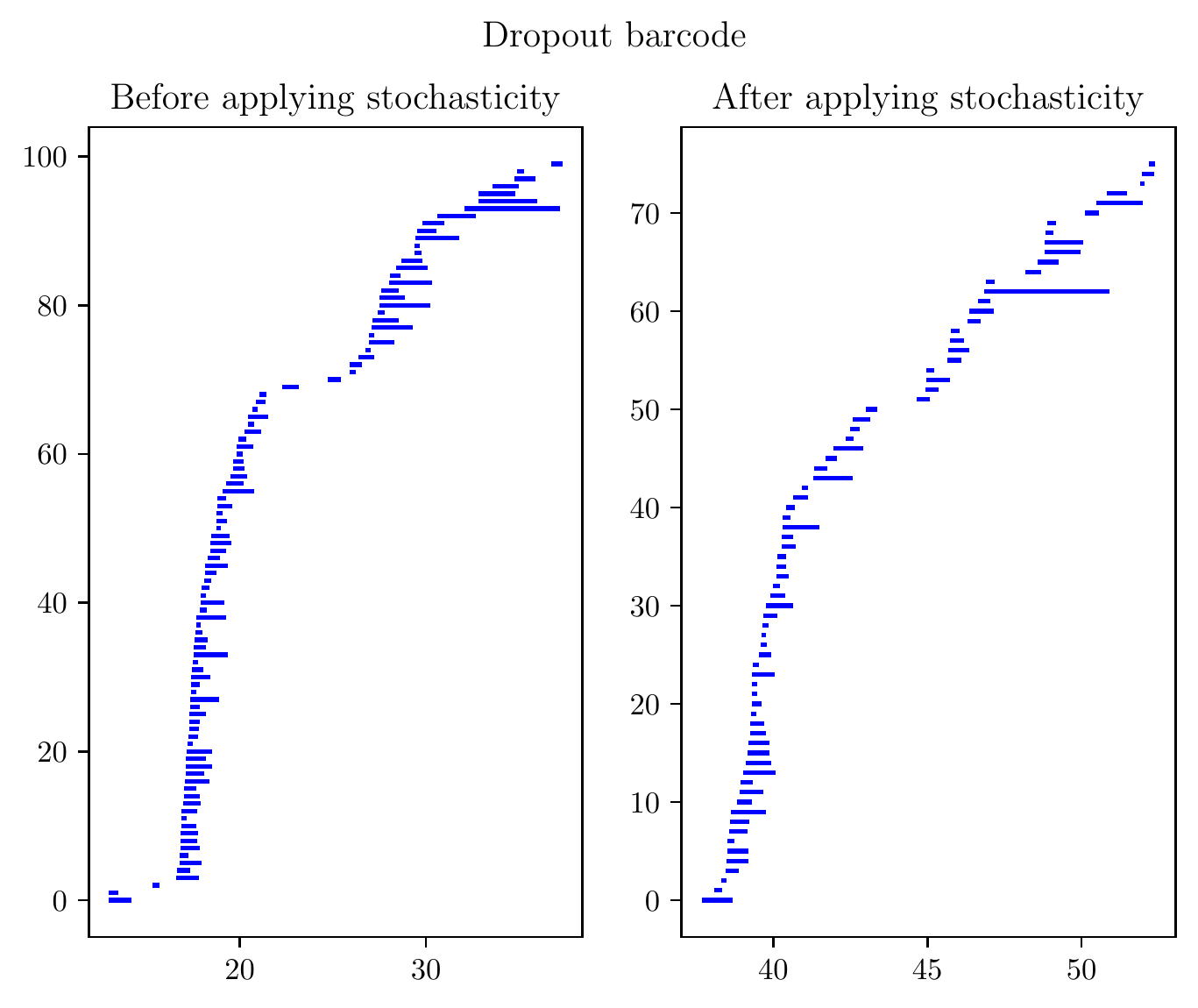}
        \caption{}
        \label{fig:supp_topology1_dropout}
    \end{subfigure}
    \hfill 
    %\vskip\baselineskip
    \begin{subfigure}[b]{0.32\linewidth}
        \centering
        \includegraphics[width=0.8\linewidth]{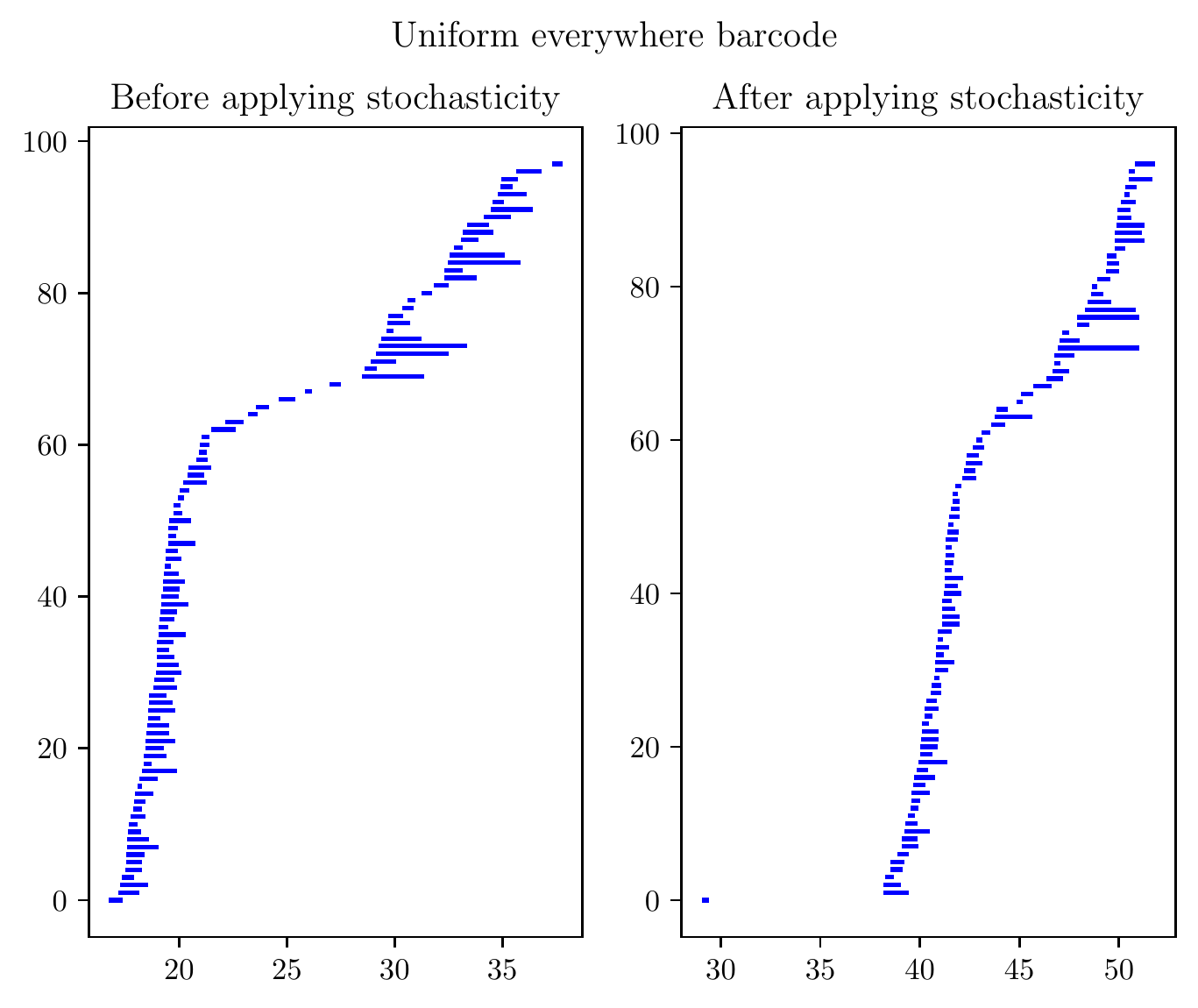}
        \caption{}
        \label{fig:supp_topology1_univorm_everywhere}
    \end{subfigure}
    \hfill    
    \begin{subfigure}[b]{0.32\linewidth}
        \centering
        \includegraphics[width=0.8\linewidth]{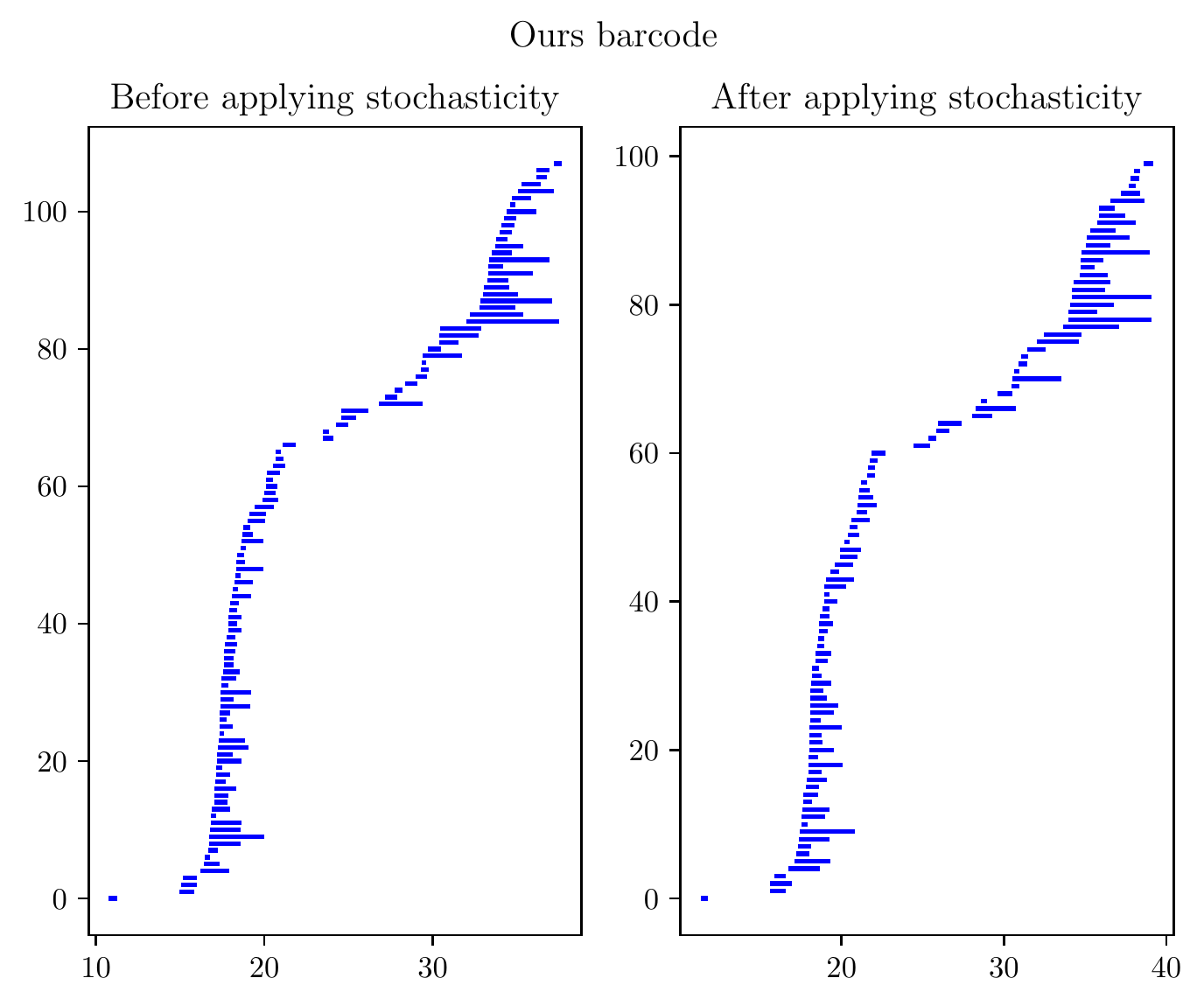}
        \caption{}
        \label{fig:supp_topology1_ours}
    \end{subfigure}
    
    \caption{\textbf{Persistence barcode diagrams before and after injecting stochasticity.} We analyze persistence barcode diagrams of feature maps before and after the stochasticity is injected. The histograms in Figures~\ref{fig:supp_topology1_bottleneck0},~\ref{fig:supp_topology1_bottleneck1}~and~\ref{fig:supp_topology1_bottleneck2} correspond to barcode distances using simplex-$0$, simplex-$1$ and simplex-$2$, respectively. The feature maps are from the last block of the transformer.  Histograms are constructed from $1000$ randomly chosen input images from the validation set. The stochasticity level is $\Delta=0.5$. In Figures~\ref{fig:supp_topology1_dropout},~\ref{fig:supp_topology1_univorm_everywhere}~and~\ref{fig:supp_topology1_ours} we can see the barcode persistence diagrams for a randomly chosen input image for dropout, non-token-consistent and token-consistent stochastic layers, respectively. Based on small distances of our method before and after injecting stochasticity, we conclude that the topological structure of token features is preserved, like it was mentioned in 
    \ifx\FileIsMerged\undefined
    Section~3 of the main paper.
    \else
    Section~\ref{sec:background} of the main paper.
    \fi
    }
    \label{fig:supp_topology_1}
\end{figure*}

%% file: figures/supplementary/topology_3.tex
\begin{figure}[t!]
    \centering
    \captionsetup{font=small}
    \begin{subfigure}[b]{0.32\linewidth}
        \centering
        \includegraphics[width=0.9\linewidth]{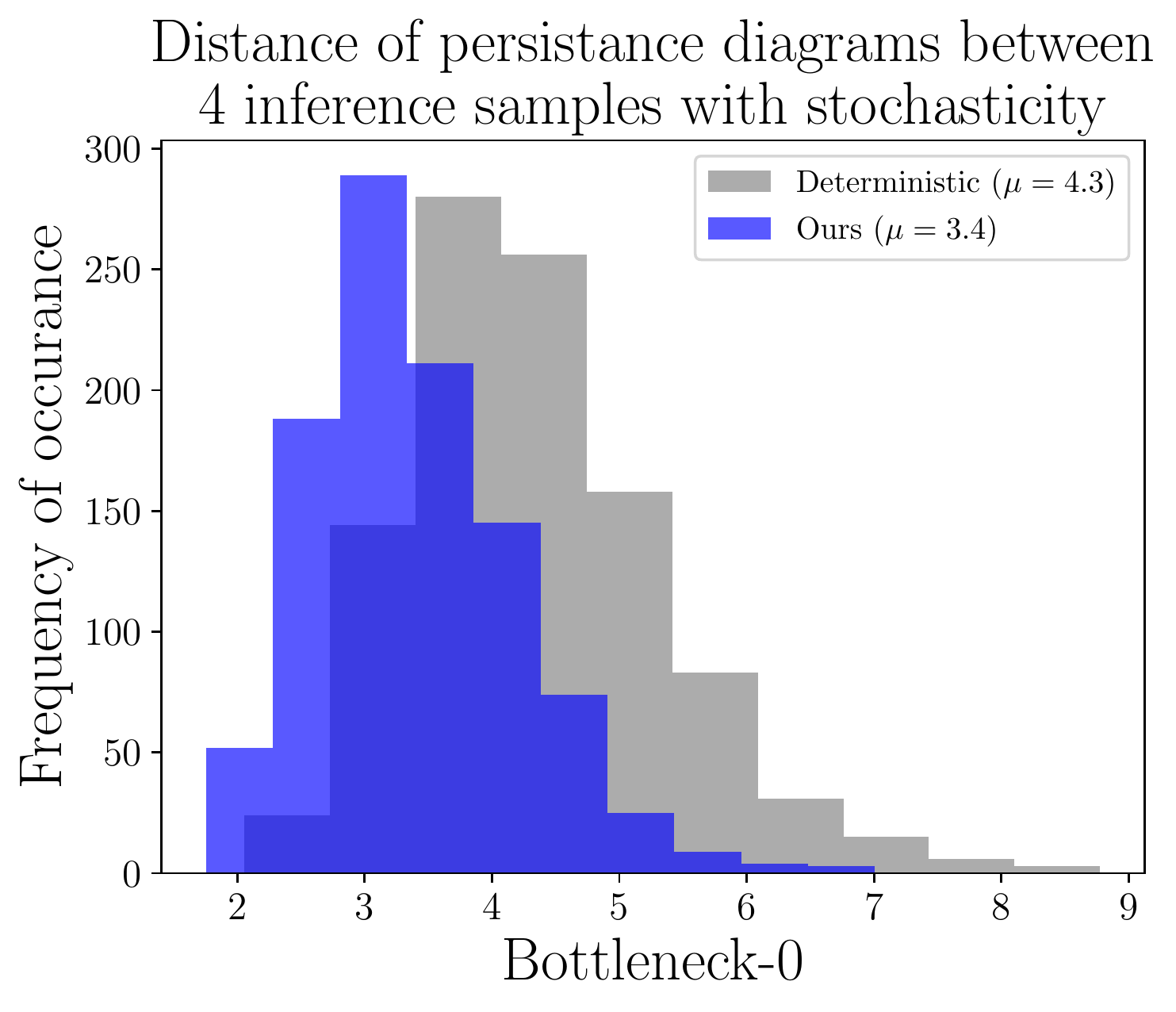}
        \caption{}
        \label{fig:supp_topology3_bottleneck0}
    \end{subfigure}
    \hfill 
    %\vskip\baselineskip
    \begin{subfigure}[b]{0.32\linewidth}
        \centering
        \includegraphics[width=0.9\linewidth]{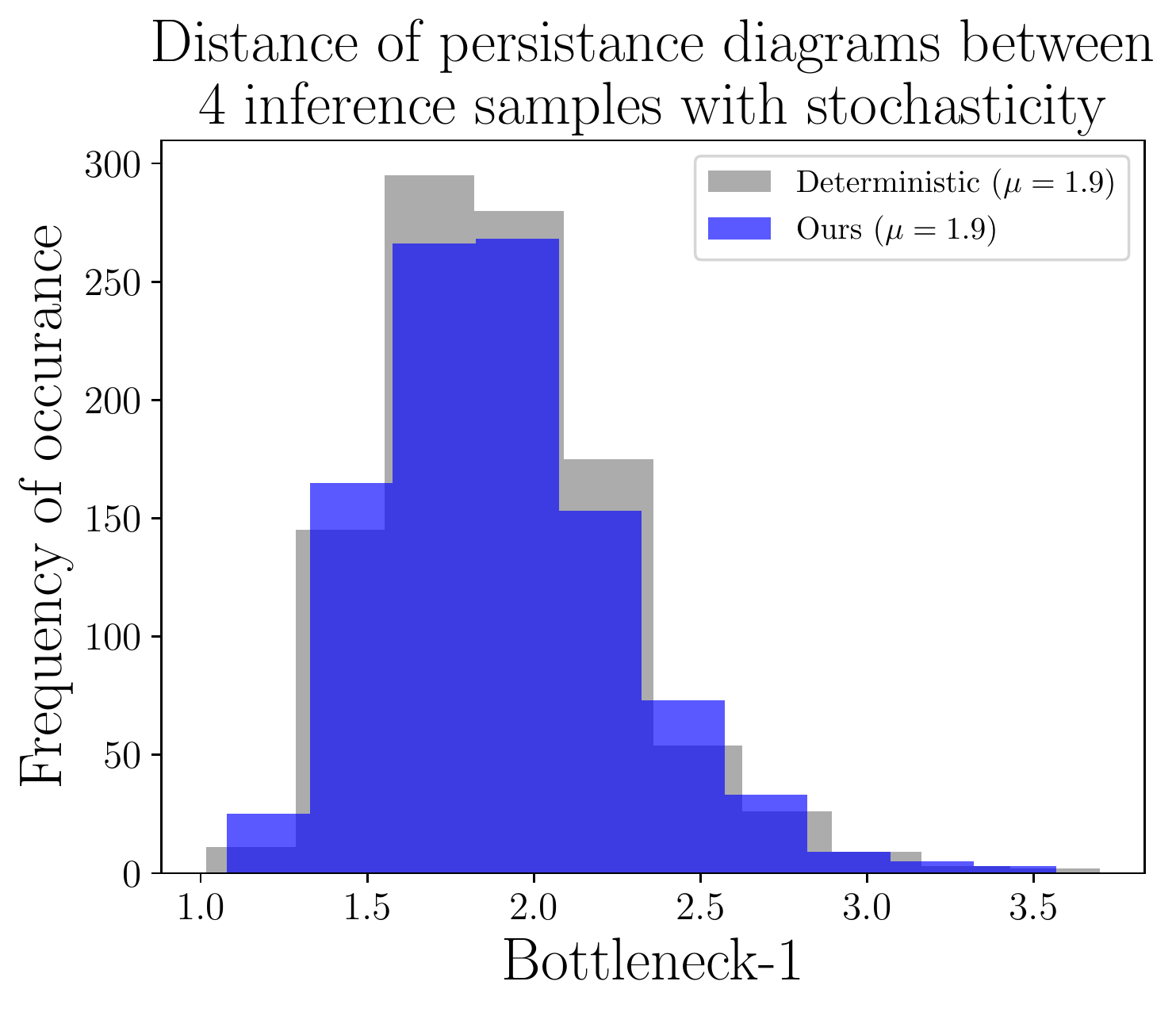}
        \caption{}
        \label{fig:supp_topology3_bottleneck1}
    \end{subfigure}
    \hfill    
    \begin{subfigure}[b]{0.32\linewidth}
        \centering
        \includegraphics[width=0.9\linewidth]{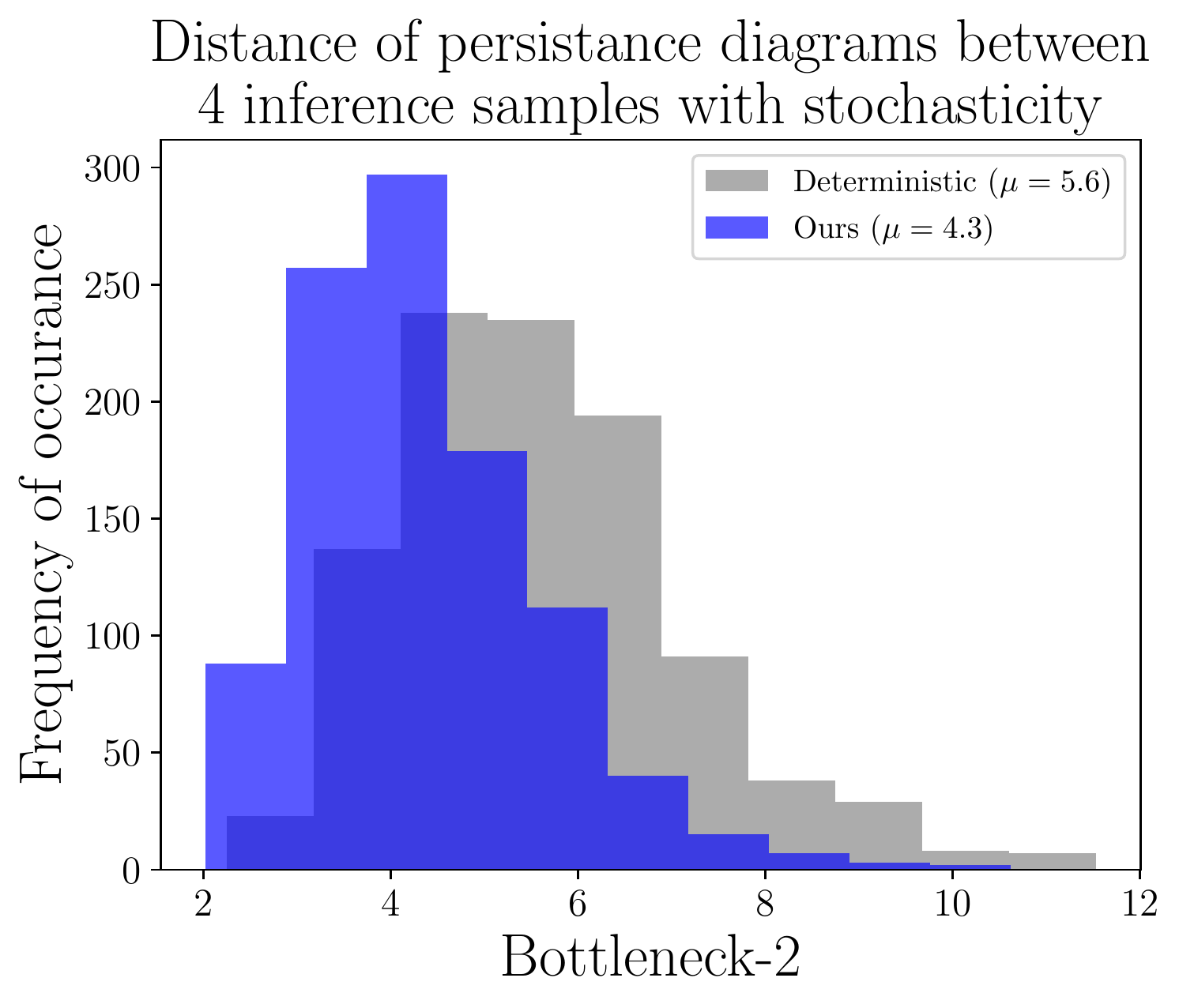}
        \caption{}
        \label{fig:supp_topology3_bottleneck2}
    \end{subfigure}

    \caption{\textbf{Distance of persistence barcode diagrams between $4$ stochasticity samples for the same input.} We see histograms of average barcode distances between $4$ stochasticity draws for the same input image. Stochasticity with the level of $\Delta=0.5$ is turned on for both the stochastic network and the deterministic baseline as well. The histograms in Figures~\ref{fig:supp_topology1_bottleneck0},~\ref{fig:supp_topology1_bottleneck1}~and~\ref{fig:supp_topology1_bottleneck2} correspond to using simplex-$0$, simplex-$1$ and simplex-$2$, respectively. We take the feature map from the last block of the transformer, right before the stochasticity is injected, and sample $4$ inference forward passes. Histograms are constructed from $1000$ randomly chosen input images from the validation set. We observe that when a stochasticity is injected, feature maps of our stochastic network have a more consistent topology (lower barcode distances) compared to the deterministic baseline. }
    \label{fig:supp_topology_3}
\end{figure}

%% file: latex/supplementary/3_discussion.tex
\section{Discussion} 
\label{sec:supp_discussion}

\subsection{Difference to standard Dropout} 
\label{sec:supp_dropout_diff}

\noindent\textbf{Theoretical.}
In its original formulation, dropout~\cite{srivastava2014dropout} aims to prevent co-adaption of neurons through randomly dropping activations, thus creating randomized neural pathways in fully-connected layers. In the case of CNNs, 2D dropout (SpatialDropout)~\cite{Tompson2015EfficientOL} has found to be a more effective regularization, due to the high correlation of nearby pixels.
In the case of vision transformers~\cite{dosovitskiy2021image} however, information of neighbouring pixels is collapsed into token features (pixels within the same input patch) as well as across tokens (pixels of different input patches). Therefore, the concept of neighbourhood gets `diluted', and the conclusions of~\cite{srivastava2014dropout} and ~\cite{Tompson2015EfficientOL} are in a state of conflict when it comes to vision transformers.
On the other hand, our token-consistent stochastic layers are designed for transformer-based architectures, and we draw our theoretical motivations from the homeomorphic nature of their operation -- by applying the same noise on all tokens
\ifx\FileIsMerged\undefined
(see Section 3. of the main paper).
\else
(see Section~\ref{sec:background} of the main paper).
\fi
Furthermore, to obtain stochastic homeomorphisms, we avoid noise distributions that violate invertibility of the operation, in particular the Bernoulli noise employed by~\cite{srivastava2014dropout,Tompson2015EfficientOL}. We therefore choose uniform noise in our experiments.

\noindent\textbf{Experimental.}
In practice, we observe the different effects of our token-consistent stochastic layers and dropout in several settings.
For instance, our token-consistent method offers better confidence calibration than dropout and the non-token consistent version, as shown in \ifx\FileIsMerged\undefined
Table~2 of the main paper.
\else
Table~\ref{table:calibration_results} of the main paper.
\fi
This pattern coincides with several observations on other tasks.
For example, in 
\ifx\FileIsMerged\undefined
Figure 4 of the main paper, 
\else
Figure~\ref{fig:collaborative_inference} of the main paper, 
\fi
it can be observed that injecting stochasticity through non-homeomorphic operations, such as dropout and the non-token-consistent coutnerpart, preserves less privacy.
Also, our token-consistent stochastic layers provide stronger stochasticity to the network under the same variance, even though the random variables are the same for each token.
This can be observed on ImageNet-1k classifcation accuracy shown in Table~\ref{table:supp_imnet_1K_accuracy}. The non-homeomorphic operations dropout and the non-token-consistent baseline tend to retain more accuracy of the original network (e.g. with $N=1$ sample).

\subsection{Fine-tuning vs. training from scratch}
\label{sec:supp_fine_tuning}
Fine-tuning brings the benefit of being able to quickly integrate our stochastic layers into an already trained network. In this process, the architecture of the original transformer is unchanged, while at the same time we can enjoy the benefits discussed in 
\ifx\FileIsMerged\undefined
Section 5 of the main paper. 
\else
Section~\ref{sec:experiments} of the main paper. 
\fi
This saves a lot of training cost and $CO_2$ emissions.
This does not significantly degrade ImageNet-1k classification accuracy as can be seen in 
\ifx\FileIsMerged\undefined
Figure 3 of the main paper.
\else
Figure~\ref{fig:classification_accuracy} of the main paper.
\fi
Moreover, we demonstrated that training the network from scratch with our stochastic layers is possible, even with strong stochasticity, in
\ifx\FileIsMerged\undefined
Table~1c in the main paper.
\else
Table~\ref{table:imnet_100_scratch_accuracy} in the main paper.
\fi

% \subsection{Privacy}
% \label{sec:supp_discuss_privacy}
% In Figure~\ref{fig:supp_reconstruction}, we show that injecting noise preserves more privacy. Reconstructed images from block $7$ of the transformer with and without our stochastic layers are shown. The first row depicts the original image. The second row depicts images reconstructed from the feature map of a deterministic network. Finally, the third row depicts reconstructions from our stochastic network. In particular the facial details are more degraded by our stochastic layers.

\subsection{Topology}
\label{sec:supp_discuss_topology}

In Figure~\ref{fig:supp_topology_1} we show histograms of barcode distances of feature maps before and after the stochasticity is injected. Based on small topological distances of features before and after injecting stochasticity with our token-consistent stochastic layers, we conclude that the topological structure of token features is preserved. This experiment complements the theory of
\ifx\FileIsMerged\undefined
Section 3 in the main paper.
\else
Section~\ref{sec:background} in the main paper.
\fi

In Figure~\ref{fig:supp_topology_3}  we see histograms of average barcode distances between $4$ stochastic samples for the same input image for our stochastic layers and the deterministic baseline.  We observe that feature maps of our stochastic network have a more consistent topology (lower barcode distances) compared to the determenistic baseline, when stochasticity is injected. 
Even though the determenistic baseline is not trained with the stochasticity, the stochasticity injected during inference preserves the topological structure of the feature map. Nevertheless, following layers damage that topological structure.
However, when the network is trained with this stochasticity source which preserves the feature map topology, following layers do less damage to the topological structure. 
%This implicates that the network learns to rely on the topological structure when making decisions.

% \subsection{Expectation over Transformation attack}
% \label{sec:supp_discuss_expectation_adversary}

% In practice, the EOT attack comes with a very high cost, since for every step of the attack one needs to compute multiple complete forward and backward passes for the same input. Furthermore, since our stochastic layers may be easily introduced to most of the popular transformer networks, the attacker can know the architecture and the weights but would still be oblivious to the presence of noise and monte-carlo sampling during inference.

% Bearing this in mind, we analyze the impact of the EOT attack with $5$ gradient samples within a $10$ step PGD with $5$ random restarts. The results are presented in Table~\ref{table:supp_adv_robustness_expectation}. Even under this strong attack, which exploits the stochasticity, we are able to retain the robustness of the regular baseline.

\subsection{Regularization effect}

It is known that stochasticity can have a regularizing effect~\cite{srivastava2014dropout}. In fact, our results in 
\ifx\FileIsMerged\undefined
Table~1c of the main paper
\else
Table~\ref{table:imnet_100_scratch_accuracy} of the main paper
\fi
hint in this direction when training from scratch. However, we did not further investigate this behaviour due to the reasons discussed in Section~\ref{sec:supp_fine_tuning} and also because it is not the main focus of this paper.

\subsection{Ethical and Societal Impact.}
This work is concerned with the learning of improved visual features by noise injection. Even though these features offer better resilience on privacy and adversarial attacks, they do not meet a certifiable standard as of now. Therefore they should not be used in a production system without additional layers of safety measures, such as encryption or humans-in-the-loop.
Although our method allows for an improved estimation of prediction confidence overall, we did not investigate the behaviour with regards to specific subgroups of classes or people. This needs to be carefully analysed with appropriate datasets to mitigate discriminatory actions by a decision-making system based on our algorithm.